\documentclass[conference]{IEEEtran}
\IEEEoverridecommandlockouts
\usepackage{cite}
\usepackage{amsmath,amssymb,amsfonts}
\usepackage{graphicx}
\usepackage{textcomp}
\usepackage{xcolor}
\usepackage{multirow}
\usepackage{cleveref}
\usepackage{algorithm}
\usepackage{algpseudocode}
\usepackage{adjustbox}
\usepackage{subcaption}
\def\BibTeX{{\rm B\kern-.05em{\sc i\kern-.025em b}\kern-.08em
    T\kern-.1667em\lower.7ex\hbox{E}\kern-.125emX}}
\begin{document}

%
%
\newcommand{\todo}[1]{{\color{red}#1}}
\newcommand{\TODO}[1]{\textbf{\color{red}[TODO: #1]}}
\newcommand{\etal}{\textit{et al.}}

\definecolor{dark-gray}{gray}{0.20}
\newcommand{\pub}[1]{{\color{dark-gray}{\tiny{[{#1}]}}}}
\newcommand{\red}[0]{\textcolor{red}}
\newcommand{\blue}[0]{\textcolor{blue}}
\newcommand{\gray}[0]{\textcolor{gray}}
\newcommand{\rot}[0]{\rotatebox{90}}

\newtheorem{definition}{Definition} 
\newtheorem{theorem}{Theorem} 
\newtheorem{theorem_supp}{Theorem} 
\newtheorem{lemma}{Lemma} 
\newtheorem{lemma_supp}{Lemma} 
\newtheorem{proof}{Proof}

\title{Invariance Principle Meets Vicinal Risk Minimization}
 
\author{
    \IEEEauthorblockN{Yaoyao Zhu$^{a,b}$, Xiuding Cai$^{a,b}$, Yingkai Wang$^{a,b}$, Dong Miao$^{a,b}$, Zhongliang Fu$^{a,b*}$, Xu Luo$^{a,b*}$}
    \IEEEauthorblockA{$^a$  Chengdu Institute of Computer Application, Chinese Academy of Sciences, Chengdu, 610213, China}
    \IEEEauthorblockA{$^b$  University of Chinese Academy of Sciences, Beijing, 101408, China}
    \IEEEauthorblockA{$^*$  Corresponding author}
    \IEEEauthorblockA{\{zhuyaoyao19, caixiuding20, wangyingkai22, miaodong20, luoxu18\}@mails.ucas.ac.cn, \{fzliang\}@casit.com.cn}}
\maketitle

\begin{abstract}
Deep learning models excel in computer vision tasks but often fail to generalize to out-of-distribution (OOD) domains. Invariant Risk Minimization (IRM) aims to address OOD generalization by learning domain-invariant features. However, IRM struggles with datasets exhibiting significant diversity shifts. 
While data augmentation methods like Mixup and Semantic Data Augmentation (SDA) enhance diversity, they risk over-augmentation and label instability.
To address these challenges, we propose a domain-shared Semantic Data Augmentation (SDA) module, a novel implementation of Variance Risk Minimization (VRM) designed to enhance dataset diversity while maintaining label consistency. 
We further provide a Rademacher complexity analysis, establishing a tighter generalization error bound compared to baseline methods. 
Extensive evaluations on OOD benchmarks, including PACS, VLCS, OfficeHome, and TerraIncognita, demonstrate consistent performance improvements over state-of-the-art domain generalization methods.
\end{abstract}


\section{Introduction}
\label{sec:intro}

\begin{figure}[t]
  \centering
  \includegraphics[width=0.35\textwidth]{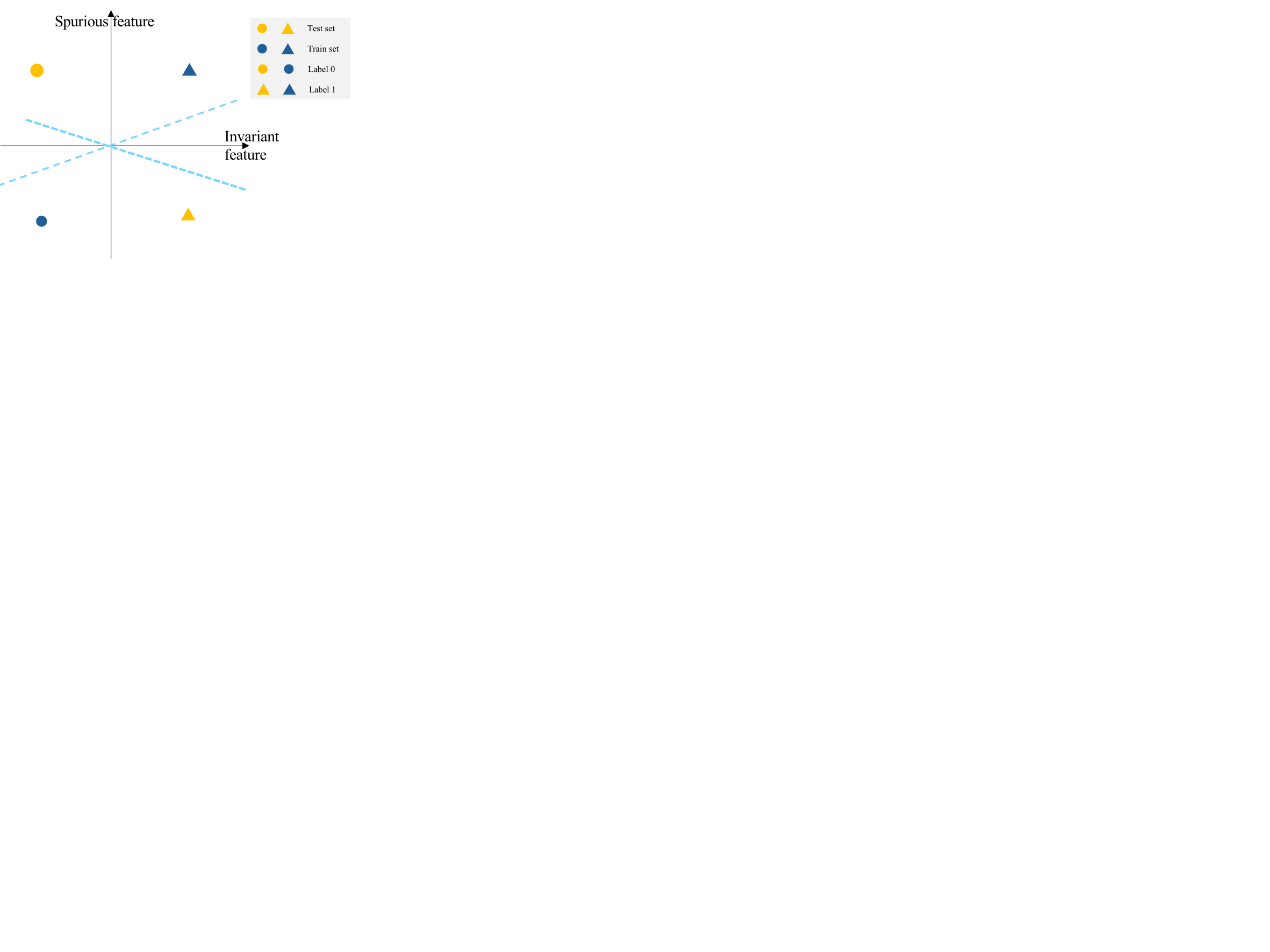}
  \caption{
  2D classification example illustrating domain generalization failed. 
  The dashed lines indicate generalization failure predictors.
 }
 \label{fig: ood_failed}
\end{figure}

Deep learning models have achieved remarkable success in computer vision~\cite{he2016deep}.
However, they frequently underperform in out-of-distribution (OOD) settings where training and test distributions diverge~\cite{gulrajani2020search, ahuja2021invariance}.
This is common in real-world applications, such as medical image analysis, where models trained on data from a single hospital struggle to generalize across datasets from other institutions~\cite{dgdr}.
Arjovsky et al.~\cite{arjovsky2019invariant} attributed the generalization failures in OOD settings to spurious correlations learned during model training.
For example, a model trained on “cows in the grass” and “camels in the sand” may misclassify “cows in the sand” or “camels in the grass,” as it relies on the background rather than object features for classification.
As illustrated in \cref{fig: ood_failed}, the background (spurious feature) and animal shape (invariant feature) are crucial to understanding the generalization failure.
The model captures false causality and uses the spurious feature as a classification criterion, failing generalization. 
This limitation significantly restricts the applicability of deep learning systems in real-world deployments.

\begin{figure*}[t]
  \centering
  \includegraphics[width=1\textwidth]{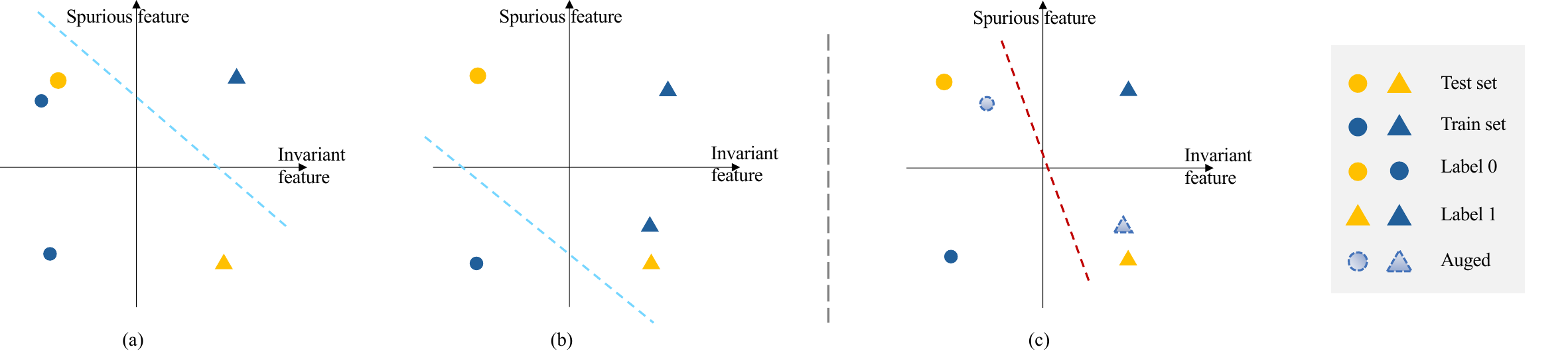}
  \caption{
    (a) and (b) Example of generalized failure in 2D classification when features do not satisfy support overlap leading to model generalization failure;
    (c)  Example of generalized success in 2D classification using semantic data augmentation to generate new samples.
 }
  \label{fig: overview} 
\end{figure*}

Invariant Risk Minimization (IRM)~\cite{arjovsky2019invariant} and its variants~\cite{ahuja2021invariance,lin2022birm,krueger2021out,chen2022pair} have emerged as crucial methods for domain generalization (DG). 
These methods aim to learn invariant features across environments, thus mitigating spurious correlations that commonly arise in empirical risk minimization (ERM)~\cite{sain1996nature}.
While IRM shows promising potential, its practical applicability is hampered by several limitations, particularly in scenarios involving high dataset diversity, which can lead to performance degradation in out-of-distribution (OOD) settings.
Ahuja \etal~\cite{ahuja2021invariance} noted that ERM and IRM may fail to address the OOD problem if features insufficiency support overlap for linear classification tasks.
Meanwhile, Elan \etal~\cite{rosenfeld2021the} highlight that IRM may encounter substantial limitations in non-linear settings unless the test data closely resemble the training distribution.
To understand the reasons for the generalization failure, Ye \etal~\cite{ye2022ood} attempt to quantify dataset shifts in terms of diversity and correlation.
The results show that IRM and its variant algorithm (VREx)~\cite{krueger2021out} outperform ERM on datasets with significant correlation shifts. 
Unfortunately, the IRM series algorithms perform poorly on datasets with significant diversity shifts.
\emph{All of this evidence points to dataset diversity shifts as the reason for the failure of IRM generalization.}
For instance, revisiting the example of 'cows in the grass' and 'camels in the sand,' even if 'cows in the desert' are added to the training set, the model may still learn spurious associations.
As illustrated in subfigure (a) of \cref{fig: overview}. 
Similarly, when there are "camels in the grass" in the training set (subfigure (b) in \cref{fig: overview} shown "cow in the sand"), the model fails to generalize.

Vicinal Risk Minimization (VRM)~\cite{chapelle2000vicinal} seeks to improve generalization by minimizing the risk over a vicinal distribution.
Mixup~\cite{zhang2017mixup} is an implementation of VRM, which generates new samples by applying transformations to the training data, effectively increasing the diversity of the dataset and enhancing model robustness.
Moreover, the Mixup series method has been widely used for domain generalization~\cite{lee_decompose_2023, zhou2021domain}.
However, mixing samples may lead to label inconsistency~\cite{concutmix}.
Recent advancements in domain generalization (DG) have explored semantic data augmentation (SDA), or feature augmentation~\cite{li2023cross}, to improve model robustness across diverse domains~\cite{li_simple_nodate, wang2024inter}. 
We can achieve semantic transformations by changing deep features, such as altering background or object textures, to increase dataset diversity and computational efficiency.
The motivation behind these operations is that deep features are typically linear~\cite{bengio2013better, upchurch2017deep}. 
Unfortunately, these approaches encounter Mixup-based methods, particularly the risk of label inconsistency due to over-augmentation~\cite{zhu2024bayesian}.
A vivid, albeit unrealistic, example is that we can only generate images of cows on Mars once it becomes a reality.
Bayesian random semantic data augmentation (BSDA)~\cite{zhu2024bayesian} tries to address this issue by estimating augmentable distribution using variational inference. 
While BSDA has shown promise in medical image classification, its potential for broader domain generalization tasks still needs to be explored.
Moreover, akin to Mixup, generating new samples through feature alteration in SDA aligns with the Vicinal Risk Minimization (VRM) principles ~\cite{chapelle2000vicinal}.

Intuitively, VRM is promising for addressing the generalization failure of IRM under conditions of substantial dataset diversity shift.
Therefore, we advocate using VRM to expand the feature overlap between domains (low overlap is a phenomenon in the feature space of dataset diversity shifts) to improve the performance of IRM.
For example, as shown in subfigure (c) of \cref{fig: overview}, the model can generalize well when the features satisfy overlap.
Unfortunately, mainstream VRM implementations for DG face the problem of label inconsistency.
At the same time, our goal is to increase the inter-domain overlap (which implies a decrease in the inter-domain dataset diversity shifts) for IRM references to VRMs, and the current BSDA may not be capable of this operation.
To address this problem, we propose a domain-shared semantic data augmentation module that expands the inter-domain feature overlap and improves the generalization ability of IRM in large dataset diversity shifts scenarios.
Specifically, we define an implementation of VRM in deep feature space in the form of feature translation. 
Unlike other VRM implementations such as Mixup~\cite{zhang2017mixup}, our method generates samples with labels consistent with the reference labels, i.e., label consistency. 
Receiving inspiration from BSDA~\cite{zhu2024bayesian}, we use variational inference to estimate the vicinal distribution and adopt from it to obtain the panning amount for semantic data augmentation.
Despite our method's simplicity, our experiments demonstrate that it effectively mitigates IRM's generalization failure while visualizing features showing an increase in overlapping regions.
The main contributions of this paper are:
\begin{itemize}
  \item We design a domain-shared semantic augmentation module, enabling controlled semantic exchanges between domains while maintaining label consistency.
  \item We derive a theoretically tighter upper bound on its generalization error relative to the baseline, leveraging Rademacher Complexity as a metric.
  \item Extensive experiments on challenging OOD benchmarks, including PACS, VLCS, OfficeHome, and TerraIncognita, consistently demonstrate that our method outperforms state-of-the-art DG models.
\end{itemize}

\section{Preliminary}
\label{sec:preliminary}




\subsection{Invariant Risk Minimization}
\label{method: irm}
IRM\cite{arjovsky2019invariant} seek to achieve the best classifier over source domains, emphasizing cross-domain generalization capability,
 which is different from the traditional ERM\cite{sain1996nature} focuses only on optimization performance on a single training set.

\begin{definition}[IRM]
	When the classifier $\mathbf{w}:\mathcal{H} \rightarrow \mathcal{Y}$ is simultaneously optimal for all domains, i.e., 
    $\mathbf{w} \in {\underset{\bar{w}:\mathcal{H} \rightarrow \mathcal{Y}} { \operatorname {arg \, min}} \, \ell_e(\bar{\mathbf{w}} \circ \mathbf{\Phi}), \forall e \in \mathcal{E}}$, 
	data representation $\mathbf{\Phi}: \mathcal{X} \rightarrow \mathcal{H}$ induces an invariant classifier $\mathbf{w} \circ \mathbf{\Phi}$ across all domains $\mathcal{E}$.
\end{definition}
Intuitively, the ideal feature is a cause of $y$, and the causal mechanism should not be affected by other mechanisms. 
Hence, this feature is a domain-invariant feature.
Formally, the IRM is defined as:

\begin{equation}
	\begin{aligned}
		&\min_{\mathbf{\Phi, \mathbf{w}}}
		\sum_{e\in\mathcal{E}_{\mathrm{tr}}} \ell_e(\mathbf{w} \circ \mathbf{\Phi})\\
		&\mathrm{s.t.} \quad \mathbf{w} \in \underset{\bar{\mathbf{w}}:\mathcal{H}\to\mathcal{Y}}{\operatorname*{\arg\min}} \, 
		\ell_e(\bar{\mathbf{w}}\circ\mathbf{\Phi}),\forall e \in \mathcal{E}_{\mathrm{tr} }.
		\end{aligned}
\end{equation}

\subsection{Vicinal Risk Minimization}
\label{sec:vrm}

Vicinal Risk Minimization (VRM)~\cite{chapelle2000vicinal} is a framework designed to improve generalization by minimizing the risk over a distribution of perturbed samples drawn from the original training data. 
\begin{definition}[VRM]
Given data representation $\mathbf{\Phi}: \mathcal{X} \rightarrow \mathcal{H}$  and a classifier $\mathbf{w}: \mathcal{H} \rightarrow \mathcal{Y}$, 
the VRM objective minimizes the expected loss over a distribution of perturbed data points sampled from the training data, as defined below:
\begin{equation}
	\begin{aligned}
		&\min_{\mathbf{\Phi, \mathbf{w}}}
		\sum_{e \in  \mathcal{E}_{\mathrm{tr}}} \sum_{\mathbf{x}_i \in e} \ell_e(\mathbf{w} \circ \mathbf{\Phi}(\tilde{\mathbf{x}}_i), \tilde{y}_i),
		\end{aligned}
\end{equation}
where $(\tilde{\mathbf{x}}_i, \tilde{y}_i)$ is vicinity of $(\mathbf{x}_i, y_i)$ sampled from vicinity distribution $v$.
\end{definition}


\section{Method}
\label{sec:method}



\subsection{VIRM}
\label{sec:virm}
As outlined in the introduction,  we advocate the use of VRM to expand the feature overlap between domains  to improve the performance of IRM, named Vicinal Invariant Risk Minimization (VIRM).
VIRM assumes that for a given feature $\mathbf{z}_i = \mathbf{\Phi}(\mathbf{x}_i)$ there exists a distribution $p(\mathbf{\xi}|\mathbf{z}_i)$ 
such that $\tilde{\mathbf{z}}_i =\mathbf{z}_i + \mathbf{\xi}_j, \forall \mathbf{\xi}_j \in p(\mathbf{\xi}|\mathbf{z}_i)$ with label $\tilde{y}_i$.
Meanwhile, we differ from some VRM implementations such Mixup~\cite{zhang2017mixup} in that we constrain $\tilde{y}_i = y_i$.
VIRM is formulated as the following optimization problem:

\begin{definition}[VIRM]
	\label{def:virm}
	Given data representation $\mathbf{\Phi}: \mathcal{X} \rightarrow \mathcal{H}$, classifier $\mathbf{w}: \mathcal{H} \rightarrow \mathcal{Y}$,
	the VIRM objective minimizes the expected loss, as defined below:
	\begin{equation}
		\begin{aligned}
			&\min_{\mathbf{\Phi, \mathbf{w}}}
			\sum_{e \in  \mathcal{E}_{\mathrm{tr}}} \sum_{\mathbf{x}_i \in e} \ell_e(\mathbf{w} (\mathbf{z}_i + \mathbf{\xi}), y_i) \\
			&\mathrm{s.t.} \quad \mathbf{w} \in \underset{\bar{\mathbf{w}}:\mathcal{H}\to\mathcal{Y}}{\operatorname*{\arg\min}} \, \ell_e(\bar{\mathbf{w}}\circ\mathbf{\Phi}),\forall e \in \mathcal{E}_{\mathrm{tr} },
			\end{aligned}
	\end{equation}
	where $\mathbf{\xi}$ is the translation magnitude sampled from the augmentable distribution $p(\mathbf{\xi} | \mathbf{z}_i)$.
\end{definition}

\subsection{Domain-Shared SDA module}
\label{sec:ds_sda}
In the VRM implementation, we need to estimate the augmentable distribution $p_e(\mathbf{\xi} |\mathbf{z}_i)$ while ensuring that $\tilde{y}_i = y_i$.
Inspired by BSDA~\cite{zhu2024bayesian}, we introduce model $q_{\phi_e}(\mathbf{\xi}|\mathbf{\mathbf{z}}_i)$ to approximate the distribution $p_e(\mathbf{\xi} |\mathbf{z}_i)$ of domain $e$.
The Kullback-Leibler (KL) divergence measures the similarity between these two distributions, aiming to make $q_{\phi_e}(\mathbf{\xi} |\mathbf{z}_i)$ closely match $p_e(\mathbf{\xi} |\mathbf{z}_i)$ by maximizing the KL divergence.
Thus, our optimization goal of the SDA module is defined as:

\begin{equation}
	\label{eq:bsda}
	\phi_e = \underset{\phi_e}{\arg\max} D_{KL}(q_{\phi_e}(\mathbf{\xi}|\mathbf{\mathbf{z}}_i) || p_e(\mathbf{\xi}|\mathbf{\mathbf{z}}_i)).
\end{equation}
Another challenge with \Cref{eq:bsda} arises when domain-specific estimators are introduced, as they prevent feature sharing across domains, hindering the transformation of domain-invariant features. 
Although using shared estimators across domains can address this issue, it also increases the optimization complexity of the model. 
This trade-off will be further explored in our subsequent experiments (\Cref{exp: ablation_study}). 
Thus, our domain-shared SDA module is defined as follows:
\begin{equation}
	\label{eq:sdsda}
	\phi = \underset{\phi}{\arg\max} D_{KL}(q_{\phi}(\mathbf{\xi}|\mathbf{\mathbf{z}}_i) || p(\mathbf{\xi}|\mathbf{\mathbf{z}}_i)).
\end{equation}

While \Cref{eq:sdsda} allows us to estimate the distribution  $p(\mathbf{\xi}|\mathbf{\mathbf{z}}_i)$, additional constraints on the augmented features are essential to maintain label consistency.
Specifically, consistency constrains as follows:
\begin{equation}
	\label{eq:label_consistency}
	\min_{\mathbf{\Phi, \mathbf{w}}, \mathbf{\phi}} \sum_{\mathbf{x}_i \in \mathcal{E}}  \ell (\mathbf{w}(\tilde{\mathbf{z}}_i), y_i).
\end{equation}

Moreover, it is impractical to enforce robustness on all features simultaneously within the IRM framework, as our objective is to identify a subset of robust features rather than making every feature invariant. 
Therefore, $\tilde{\mathbf{z}}_i$ is generated using the following equation
\begin{equation}
	\tilde{\mathbf{z}} = \mathbf{z} + \mathbf{d}_{\lambda} \odot \mathbf{\xi},
	\label{eq:z_tilde}
\end{equation}
where $d_i \in \mathbf{d}_{\lambda} \sim \text{Bernoulli}(\lambda),\, \forall i = 1, 2, \ldots, k$ and $\mathbf{d}$ is $k$ dimension.
The role of $\mathbf{d}$ is to be used to select features for panning, and it is difficult for us to quantify spurious features and invariant features in the framework of IRM. 
Therefore, in our approach, we treat every feature equally. 
This is motivated by Ahuja \etal.~\cite{ahuja2021invariance} pointing out that invariant feature overlap is a necessary condition for the success of the generalization, while at the same time if spurious invariant feature overlap is violated, the IRM fails to generalize.

\subsection{Loss Function}
\label{sec:loss}
Our method is formulated as the following optimization problem:
\begin{equation}
	\label{eq:virm}
	\begin{aligned}
		\mathcal{L}_{\text{VIRM}}(\Phi, \mathbf{w}, \phi) 
		& = \mathcal{L}_{\text{IRM}} + \alpha \mathcal{L}_{\phi}.
	\end{aligned}
\end{equation}

\Cref{eq:virm} consists of two components, which are the invariant risk and the loss function of the module of domain-shared SDA module $\mathcal{L}_{\phi}$.
The loss function of the domain-shared SDA module is defined as follows:
\begin{equation}
	\begin{aligned}
	& \mathcal{L}_{\phi} = 
	& - \frac{1}{2}\sum_{i=0}^n(1 + \log(\boldsymbol{\sigma}^2) - \boldsymbol{\sigma}^2 ) + 
	\frac{1}{2n}\sum_{l=1}^n (\mathbf{\hat{z}_i} - \mathbf{z}_i)^2.
  \label{eq:loss_dssda}
  \end{aligned}
\end{equation}
For more details about \Cref{eq:loss_dssda}, please refer to \Cref{sec:detial_loss}.

For first term $\mathcal{L}_{\text{IRM}}$ in \Cref{eq:virm}, there are various approaches to implementing IRM, and our experiments demonstrate that combining VRM with the VREx~\cite{krueger2021out} method yields superior performance. 
For detailed analysis, please refer to \cref{exp: ablation_study}. 
The optimization objectives of VREx are defined as follows:
\begin{equation}
	\label{eq:vrex}
	\mathcal{L}_{\text{VREx}}(\Phi, \mathbf{w}) = \min_{\Phi, \mathbf{w}}  \beta \cdot \text{Var}\left(\{\ell_e(\Phi, \mathbf{w})\}_{e=1}^{M}\right) + \sum_{e=1}^{M} \ell_e(\Phi, \mathbf{w}),
\end{equation}

where $ \theta $ represents the model parameters, $\ell_e(\theta) $ denotes the loss on the $e$-th training domain, $ M$ is the number of training domains, $ \beta $ is a positive weight, and $ \text{Var} $ indicates the variance.

\begin{table*}[t]
	\renewcommand\arraystretch{1.0}
	\centering
	\caption{
		Evaluations on the DomainBed benchmark with leave-one-domain-out cross-validation model selection. 
		The best results are highlighted in \textcolor{red}{red}, and the second-best results are highlighted in \textcolor{blue}{blue}.
	}
	\begin{tabular}{l|ccccc}
	  \hline
	\textbf{Method} & \textbf{PACS} & \textbf{VLCS} & \textbf{OfficeHome} & \textbf{TerraInc} & \textbf{Avg.} \\ 
  \hline
	VNE~\pub{CVPR'23}~\cite{kim2023vne}  & 80.5 $\pm$  0.2 & \blue{76.7 $\pm$  0.2} & 58.1 $\pm$ 0.1 & 42.5 $\pm$ 0.3 & 64.5 \\
	HYPO~\pub{ICLR'24}~\cite{bai2024hypo} & 78.9 & 76.0 & 62.2 & 36.3 & 63.4 \\
DecAug~\pub{AAAI'21}~\cite{bai2021decaug} & 82.4 & 74.6 & 62.8 & 43.1 & 65.7 \\
ERM~\cite{sain1996nature} & 79.8 $\pm$ 0.4 & 75.8 $\pm$ 0.2 & 60.6 $\pm$ 0.2 & 38.8 $\pm$ 1.0 & 63.8 \\
IRM~\cite{arjovsky2019invariant} & 80.9 $\pm$ 0.5 & 75.1 $\pm$ 0.1 & 58.0 $\pm$ 0.1 & 38.4 $\pm$ 0.9 & 63.1 \\
VREx~\pub{ICML'21}~\cite{krueger2021out} & 80.2 $\pm$ 0.5 & 75.3 $\pm$ 0.6 & 59.5 $\pm$ 0.1 & 43.2 $\pm$ 0.3 & 64.6 \\
MMD~\pub{CVPR'18}~\cite{li2018domain} & 81.3 $\pm$ 0.8 & 74.9 $\pm$ 0.5 & 59.9 $\pm$ 0.4 & 42.0 $\pm$ 1.0 & 64.5 \\
RSC~\pub{ECCV'20}~\cite{huang2020self} & 80.5 $\pm$ 0.2 & 75.4 $\pm$ 0.3 & 58.4 $\pm$ 0.6 & 39.4 $\pm$ 1.3 & 63.4 \\
ARM~\cite{zhang2020adaptive} & 80.6 $\pm$ 0.5 & 75.9 $\pm$ 0.3 & 59.6 $\pm$ 0.3 & 37.4 $\pm$ 1.9 & 63.4 \\
DANN~\pub{JMLR'16}~\cite{ganin2016domain} & 79.2 $\pm$ 0.3 & 76.3 $\pm$ 0.2 & 59.5 $\pm$ 0.5 & 37.9 $\pm$ 0.9 & 63.2 \\
GroupGRO~\pub{ICLR'20}~\cite{sagawa2019distributionally} & 80.7 $\pm$ 0.4 & 75.4 $\pm$ 1.0 & 60.6 $\pm$ 0.3 & 41.5 $\pm$ 2.0 & 64.6 \\
CDANN~\pub{ECCV'18}~\cite{li2018deep} & 80.3 $\pm$ 0.5 & 76.0 $\pm$ 0.5 & 59.3 $\pm$ 0.4 & 38.6 $\pm$ 2.3 & 63.6 \\
CAD~\pub{ICLR'22}~\cite{ruan2021optimal} & 81.9 $\pm$ 0.3 & 75.2 $\pm$ 0.6 & 60.5 $\pm$ 0.3 & 40.5 $\pm$ 0.4 & 64.5 \\
CondCAD~\pub{ICLR'22}~\cite{ruan2021optimal} & 80.8 $\pm$ 0.5 & 76.1 $\pm$ 0.3 & 61.0 $\pm$ 0.4 & 39.7 $\pm$ 0.4 & 64.4 \\
MTL~\pub{JMLR'21}~\cite{blanchard2021domain} & 80.1 $\pm$ 0.8 & 75.2 $\pm$ 0.3 & 59.9 $\pm$ 0.5 & 40.4 $\pm$ 1.0 & 63.9 \\
MixStyle~\pub{ICLR'21}~\cite{zhou2021domain} & 82.6 $\pm$ 0.4 & 75.2 $\pm$ 0.7 & 59.6 $\pm$ 0.8 & 40.9 $\pm$ 1.1 & 64.6 \\
MLDG~\pub{AAAI'18}~\cite{li2018learning} & 81.3 $\pm$ 0.2 & 75.2 $\pm$ 0.3 & 60.9 $\pm$ 0.2 & 40.1 $\pm$ 0.9 & 64.4 \\
Mixup~\cite{yan2020improve} & 79.2 $\pm$ 0.9 & 76.2 $\pm$ 0.3 & 61.7 $\pm$ 0.5 & 42.1 $\pm$ 0.7 & 64.8 \\
MIRO~\pub{ECCV'22}~\cite{cha2022domain} & 75.9 $\pm$ 1.4 & 76.4 $\pm$ 0.4 & \red{64.1 $\pm$ 0.4} & 41.3 $\pm$ 0.2 & 64.4 \\
Fishr~\pub{ICML'22}~\cite{rame2021ishr} & 81.3 $\pm$ 0.3 & 76.2 $\pm$ 0.3 & 60.9 $\pm$ 0.3 & 42.6 $\pm$ 1.0 & 65.2 \\
SagNet~\pub{CVPR'21}~\cite{nam2021reducing} & 81.7 $\pm$ 0.6 & 75.4 $\pm$ 0.8 & 62.5 $\pm$ 0.3 & 40.6 $\pm$ 1.5 & 65.1 \\
SelfReg~\pub{ICCV'21}~\cite{kim2021selfreg} & 81.8 $\pm$ 0.3 & 76.4 $\pm$ 0.7 & 62.4 $\pm$ 0.1 & 41.3 $\pm$ 0.3 & 65.5 \\
Fish~\pub{ICLR'21}~\cite{shi2021gradient} & 82.0 $\pm$ 0.3 & 76.9 $\pm$ 0.2 & 62.0 $\pm$ 0.6 & 40.2 $\pm$ 0.6 & 65.3 \\
CORAL~\pub{ECCV'16}~\cite{sun2016deep} & 81.7 $\pm$ 0.0 & 75.5 $\pm$ 0.4 & 62.4 $\pm$ 0.4 & 41.4 $\pm$ 0.5 & 65.3 \\
SD~\pub{NeurIPS'21}~\cite{pezeshki2021gradient} & 81.9 $\pm$ 0.3 & 75.5 $\pm$ 0.7 & 62.9 $\pm$ 0.3 & 42.0 $\pm$ 1.0 & 65.6 \\
DGRI~\pub{ICCV'23}~\cite{chen2023domain} & \blue{82.8 $\pm$ 0.3} & 75.9 $\pm$ 0.3 & 63.3 $\pm$ 0.1 & 43.7 $\pm$ 0.5 & \blue{66.4} \\
W2D~\pub{CVPR'22}~\cite{huang2022two} & 80.9 $\pm$ 0.6 & 74.4 $\pm$ 0.3 & 58.8 $\pm$ 0.3 & 35.9 $\pm$ 0.6 & 62.5 \\
	\hline
	\textbf{Ours}  & \red{83.3 $\pm$ 0.2} & \red{77.5 $\pm$ 0.2} & \blue{63.3 $\pm$ 0.0} & \red{44.3 $\pm$ 0.4} & \red{67.1}\\
	\hline
	\end{tabular}
	\label{tab:com}
  \end{table*}

\subsection{Generalization Error Bound}
\label{sec:err_bound_bsda}

The Rademacher complexity is a measure of the complexity of a function class, which can be used to derive generalization bounds for learning algorithms.

\begin{lemma}
	\label{lemma:bound}
	Let $S$ be training set, hypothesis class $\mathbb{H}$, loss function $\ell$.
	Then:
	\begin{equation}
		\label{eq:bound}
		\frac{1}{2} \mathbb{E}_S R(\bar{\mathcal{F}} \circ \mathcal{S}) 
		\le \mathbb{E}_S \sup_{f \in \mathcal{F}} |\mathbb{E} f - \hat{\mathbb{E}}f | 
		\le 2 \mathbb{E}_S R(\mathcal{F} \circ \mathcal{S}),
	\end{equation}
	where $\hat{\mathbb{E}}f$ is the empirical expectation of $f$, and to simplify the notation, let $\mathcal{F}\overset{\mathrm{def}}{\operatorname*{=}}\ell\circ\mathcal{H}\overset{\mathrm{def}}{\operatorname*{=}}\{z\mapsto\ell(h,z):h\in\mathcal{H}\}$ , $\bar{\mathcal{F}}=\{f-\mathbb{E}[f]\mid f\in\mathcal{F}\}$.
\end{lemma}

\Cref{lemma:bound} establishes a fundamental relationship between the empirical Rademacher complexity  and the uniform convergence bound on the generalization error. 

\subsubsection{Rademacher Complexity of Linear Classes}
Before analyzing the Rademacher complexity of \Cref{eq:z_tilde}, we first examine the generalization bound for linear classes as the baseline.
\begin{lemma}
	\label{lemma:rc}
	Let $S = (\mathbf{x}_1,...,\mathbf{x}_m)$ be vectors in a Hilbert space. Define: $\mathcal{H} \circ S = \{ (\langle\mathbf{w},\mathbf{x}_1\rangle, ...,\langle\mathbf{w},\mathbf{x}_m\rangle):||\mathbf{w}||_2 \le 1 \}$. 
	Then:
	\begin{equation}
			R(\mathcal{H}\circ S)~\leq~ \sqrt{\frac{\max_i\|\mathbf{x}_i\|_2^2}{m}}.
			\label{eq:rc_bs}
	\end{equation}
\end{lemma}

\begin{theorem}
	\label{theorem:bsda_rc}
	Let $\mathcal{H}$ be the hypothesis class, $S$ be the training set, and $\mathcal{H}\circ S$ be the linear class.
	Then the Rademacher complexity of the \Cref{eq:z_tilde} is bounded as follows:
	\begin{equation}
		\tilde{R}(\mathcal{H}\circ S) ~\leq~ \sqrt{ \frac{(\max_i \left\| \mathbf{x}_i  \right\|_2^2 + k)}{n} } \\
		\label{eq:rc_ours}
	\end{equation}
	where $n$ is the number of samples augmentation, is a multiple of $m$ generally, and $k$ is feature dimension.
\end{theorem}



Although the numerator includes an additional constant \(k\), in typical scenarios where \(n \gg k\), the effect of \(k\) becomes negligible.
Furthermore, since \(n > m\), the bound in \Cref{eq:rc_ours} is smaller than that in \Cref{eq:rc_bs}.
Proofs of \Cref{lemma:bound}, \Cref{lemma:rc}, and \Cref{theorem:bsda_rc} are provided in the \Cref{sec: appd_bsda_rc}.

\begin{table}[t]
  \renewcommand\arraystretch{1.2}
  \centering
  \caption{
      Average accuracy results on the ColoredMNIST dataset are reported based on the DomainBed benchmark\cite{gulrajani2020search}.
      \textbf{A} indicates the use of the domain-shared SDA module.
      \textbf{V} indicates that the original model was constrained by VREx\cite{krueger2021out}.
      \textbf{VA} indicates that the features augmented by  domain-shared SDA were subjected to VREx constraints. 
      }
      \adjustbox{max width=0.45\textwidth}{%
    \begin{tabular}{ccccccc}
    \hline
    \textbf{A} & \textbf{V} & \textbf{VA}  & \textbf{$+$90\%}    & \textbf{$+$80\%}    & \textbf{$-$90\%}    & \textbf{Avg.} \\
    \hline
    & &     & 50.0   & 50.1    & 10.0   & 36.7   \\
    \checkmark & &   & 49.7    & 54.8   & 9.8  & 38.1 \\
    & \checkmark & & 50.2    & 50.5    & \blue{10.1 }  & 36.9  \\
    \checkmark & & \checkmark  & 50.2       & \blue{57.6 }       & 10.1   & \blue{39.3} \\
    \checkmark & \checkmark & \checkmark & \blue{50.2 }  & 50.1  & \red{10.2 }  & 36.9  \\
      \hline
    \checkmark & \checkmark &  & \red{50.6 }  & \red{59.7 }  & \blue{10.1}  & \red{40.1}    \\
    \hline
  \end{tabular}}
  \label{tab:abl_st}
  \end{table}

\section{Experiment}
\label{sec:experiment}

\subsection{DomainBed Benchmark}
DomainBed~\cite{gulrajani2020search} is a widely used benchmark for DG, offering implementations of various baseline methods.
For our experiments, we use four widely recognized domain generalization benchmark datasets: PACS~\cite{pacs}, VLCS~\cite{vlcs}, OfficeHome~\cite{officehome}, and TerraIncognita~\cite{beery2018recognition}.

\textbf{Dataset:}
The PACS~\cite{pacs}, VLCS~\cite{vlcs}, OfficeHome~\cite{officehome}, and Terra Incognita~\cite{beery2018recognition} datasets consist of 9,991, 10,729, 15,588, and 24,788 examples, respectively, each with 4 domains and dimensions of (3, 224, 224), covering 7, 5, 65, and 10 classes.

\textbf{Evaluation Protocol}
We use leave-one-domain-out cross-validation for evaluation. Given $e_{\text{tr}}$ training domains, we train $e_{\text{tr}}$ models with identical hyperparameters, each leaving out one domain for validation. 
This approach assumes that domains are drawn from a meta-distribution, aiming to maximize expected performance.

\begin{figure}[t]
  \centering
  \caption{
    Kernel density estimation of feature distributions across domains on the VLCS dataset (Category: Car).
  }
  \begin{subfigure}[b]{0.23\textwidth}
      \centering
      \includegraphics[width=\textwidth]{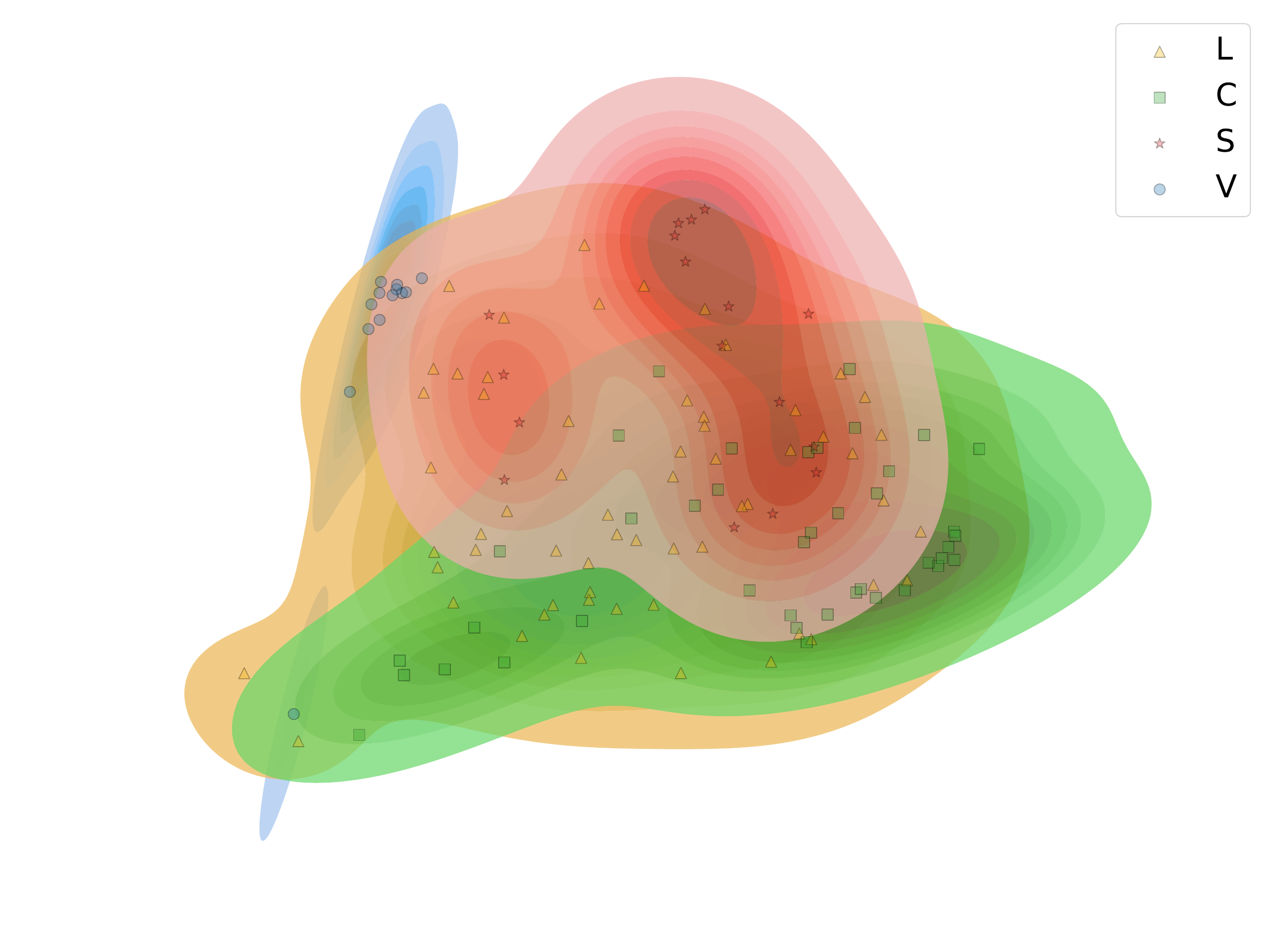}
      \caption{Original features}
      \label{fig: umap_original_overlap}
  \end{subfigure}
  \hfill
  \begin{subfigure}[b]{0.23\textwidth}
      \centering
      \includegraphics[width=\textwidth]{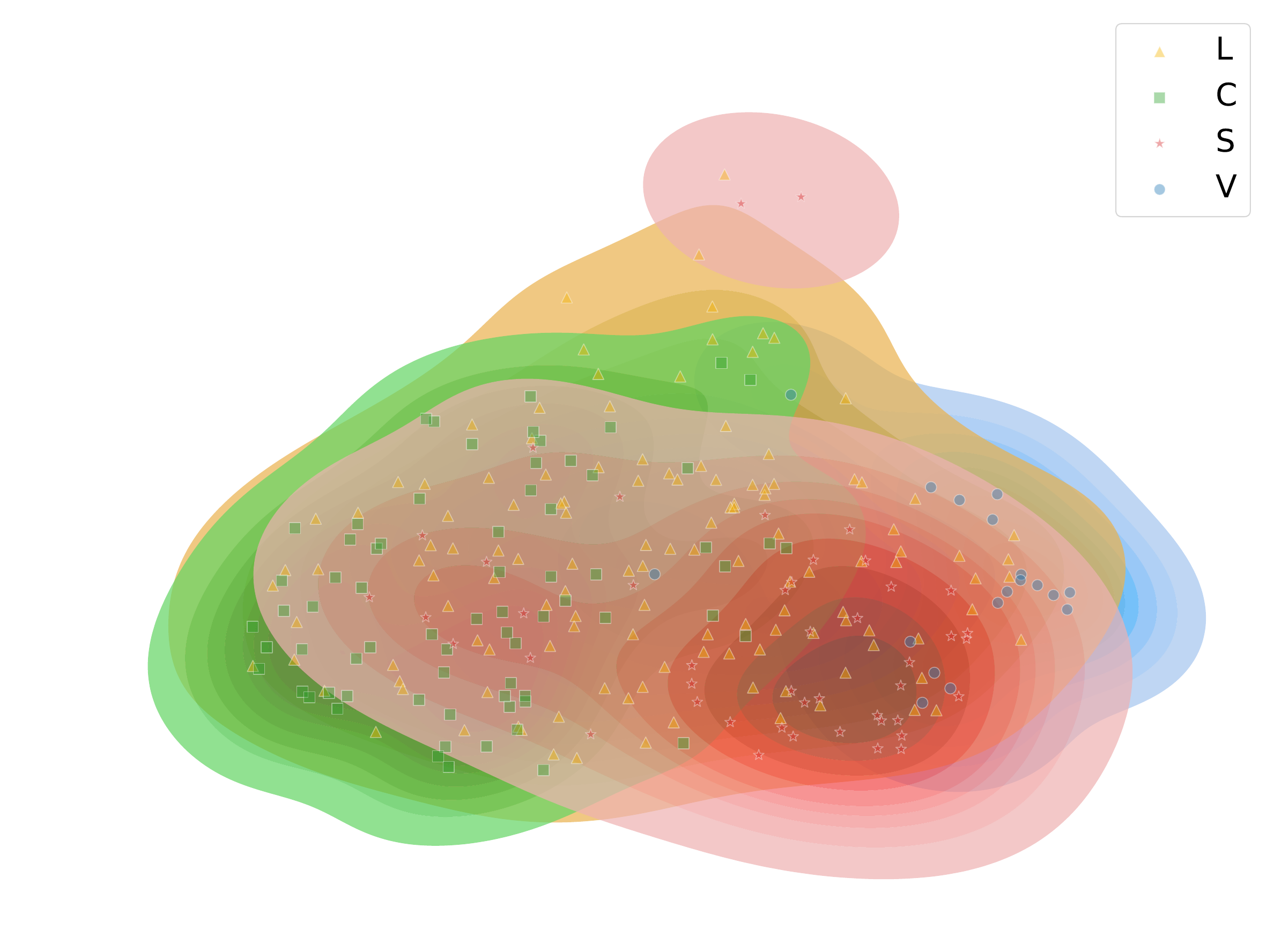}
      \caption{Augmented features}
      \label{fig: umap_aug_overlap}
  \end{subfigure}
  \label{fig: vis_umap_kde}
\end{figure}

\textbf{Experimental setting:}
We use ResNet-18~\cite{he2016deep}, pre-trained on ImageNet~\cite{deng2009imagenet}, as the feature extractor.
The Adam optimizer~\cite{kingma2014adam} is used with hyperparameters from DomainBed.
Training is conducted with PyTorch 2.0.1 on Ubuntu 22.04, using two NVIDIA GeForce RTX 4090 GPUs and an Intel Core i9-14900K CPU.

\textbf{Implementation details of domain-shared SDA:}
The encoder in domain-shared SDA uses a two-layer MLP with nodes matching the feature dimension, GeLU activation, and batch normalization before activation.
The decoder follows the same structure.
Domain-shared SDA introduces three key hyperparameters: augmentation direction probability $\lambda=0.8$, augmentation multiplicity $U=10$, and loss weight $\alpha=0.5$.
Other hyperparameters follow DomainBed’s default settings.

\subsection{Comparison Experiment}
\label{exp: domainbed}
We compare our method against state-of-the-art domain generalization algorithms. 
It is important to note that VNE~\cite{kim2023vne}, DecAug~\cite{bai2021decaug}, and HYPO~\cite{bai2024hypo} are not based on DomainBed benchmark implementations, but since their experimental setups are similar, we have included their results for comparison.

\textbf{Experimental results:}
\Cref{tab:com} shows the accuracy results on the DomainBed benchmark. 
Most of the results for other methods are taken from Chen et al.~\cite{chen2023domain}, who used the same testing platform, network architecture, and dataset. 
We reproduced the experimental results for four methods—W2D, VNE, DecAug, and HYPO—under the same experimental conditions.
The results indicate that the our method consistently outperforms existing popular DG methods.

\subsection{Ablation Study}
\label{exp: ablation_study}

This subsection examines three implementations of IRM, with the results presented in \cref{tab:abl_st}. 
The results indicate optimal performance when applying VREx constraints solely to the original features. 
A comparison of the first and third rows reveals minimal improvement in network performance with the addition of further VREx constraints. 
Moreover, applying VREx constraints simultaneously to both original and augmented features tends to reduce network performance.
The SDA module can be seen as a regularization approach that constrains the original feature representation, improving robustness but potentially limiting domain generalization. 
The domain-shared semantic data enhancement aims to amplify specific features within the training domain, thus facilitating faster and more straightforward classification. 
However, this approach may inadvertently conflict with the objectives of Invariant Risk Minimization (IRM) because these enhanced features may be domain-specific. 
Consequently, in a new domain, such features could mislead the classifier, resulting in incorrect predictions.

\subsection{Visualization}
\label{exp: vis}
To analyze feature overlap, we visualized the car category from VLCS using KDE.
\Cref{fig: umap_original_overlap} and \Cref{fig: umap_aug_overlap} demonstrate that augmentation significantly improves feature overlap, particularly in the blue region.

\section{Conclusion}
\label{sec:conclusion}

We propose a domain-shared SDA module that integrates VRM principles into IRM, offering a robust theoretical foundation via Rademacher Complexity and achieving state-of-the-art performance on OOD benchmarks.

\bibliographystyle{IEEEbib}
\bibliography{icme2025references}

\clearpage

\appendix



\subsection{Releated Work}

\subsubsection{Domain Generalization}
Domain generalization aims to generalize effectively to unseen target domains by learning from one or more source domain datasets\cite{zhou2022domain}.
Existing DG methods include domain meta-learning\cite{li2018learning}, data augmentation (DA)\cite{yan2020improve, zhou2021domain}, decoupled representations\cite{bai2021decaug}, causal reasoning\cite{sun2016deep}, 
and invariant risk minimization (IRM) series \cite{arjovsky2019invariant, krueger2021out, chang2020invariant, ahuja2021invariance}.
This section introduces invariant risk minimization and methods based on data augmentation for DG.

\subsubsection{Invariant Risk Minimization}
Arjovsky \etal~\cite{arjovsky2019invariant} proposed the IRM framework to enhance model generalization by identifying invariant features across different domains. 
BIRM~\cite{lin2022birm} applies Bayesian estimation to mitigate overfitting risks in limited training environments by accounting for uncertainty in the data, further enhancing the model’s generalization capabilities.
However, the bi-level optimization problem in IRM is complex and computationally intractable in practical scenarios. 
Many subsequent works simplify IRM by introducing constraints to transform the bi-level optimization into a more feasible single-level problem~\cite{chen2022pair,ahuja2021invariance,lin2022birm}. 
For example, PAIR~\cite{chen2022pair} addresses by framing the problem as a multi-objective optimization to achieve Pareto-optimal solutions over ERM, IRMv1, and VREx penalties, yielding more robust generalization performance. 
IRM is also highly sensitive to risk variance across training domains, especially when there is a large discrepancy between training and test distributions, a limitation that VREx~\cite{ahuja2021invariance} addresses by minimizing this variance to improve robustness against extreme distributional shifts and enhance performance in mixed causal and non-causal settings.
Lastly, IRM also needs help in ensuring sufficient invariant predictive factors, as it may inadequately capture robust features within each domain, leading to a reliance on non-invariant, environment-specific factors that harm generalization. 
IB-IRM~\cite{rosenfeld2020risks} tackles this by compressing the entropy of the feature extractor through the information bottleneck method, thus minimizing reliance on environment-specific features and improving invariant prediction. 

\subsubsection{Data Augmentation}
Other DG methods increase dataset diversity through DA~\cite{yan2020improve, zhang_domain_2024}.
Mixup~\cite{zhang2017mixup} generated new data by mixing two samples, and the Mixup series method has been widely used for domain generalization~\cite{lee_decompose_2023, zhou2021domain}.
For example, MixStyle~\cite{zhou2021domain} generates new domain style samples by mixing the styles between different domains.
Lee \etal~\cite{lee_decompose_2023} applied the Mixup technique in the frequency domain to augment the shape features.
However, these methods still have limitations in that mixing images from two classes does not necessarily result in enhanced samples belonging to them~\cite{concutmix}.

\subsubsection{Semantic Data Augmentation}
Recent work has introduced Semantic Data Augmentation (SDA), also known as feature augmentation, which applies semantic transformations to features such as background or foreground textures to increase diversity while maintaining computational efficiency~\cite{li_simple_nodate, wang2024inter}. 
For example, Li et al.\cite{li_simple_nodate} proposed adding Gaussian noise to feature embeddings during training to enhance domain generalization. In contrast, Wang et al.\cite{wang2024inter} introduced an SDA-based method to improve diversity across both intra-class and inter-domain variations.
These approaches leverage the typically linear nature of deep features~\cite{bengio2013better, upchurch2017deep}. 
However, like Mixup-based techniques, SDA needs help with label consistency due to over-augmentation, potentially leading to misalignment between the generated features and their original labels~\cite{zhu2024bayesian}. 
To address this, BSDA incorporates variational inference to control semantic transformations and ensure label consistency. 
Despite its success in medical image classification, the broader applicability of BSDA for natural image classification and domain generalization still needs to be explored.

\subsubsection{Vicinal Risk Minimization}
VRM~\cite{chapelle2000vicinal} is a framework that seeks to improve generalization by minimizing the risk over the distribution of perturbed, yet similar, data points drawn from the original training data. 
Mixup~\cite{zhang2017mixup} and its variants~\cite{zhou2021domain, lee_decompose_2023} techniques are implementations of VRM, as they generate new samples by applying transformations to the training data, effectively increasing the diversity of the dataset and enhancing model robustness. 
Similarly, SDA~\cite{ISDA} can be viewed as a form of VRM, where feature-level semantic transformations are applied to the data, increasing intra-class and inter-domain diversity~\cite{wang2024inter}.

\subsection{Dissuction}

Data Augmentation (DA) methods, such as Mixup~\cite{zhang2017mixup}, can be viewed as a special case of VRM, where new samples are generated by mixing two or more samples. 
DA techniques align with the VRM objective of improving generalization by generating data from perturbed versions of the training samples. 
Additionally, semantic data augmentation generating data by changing features, such as SFA~\cite{li2021simple} and ISDA~\cite{ISDA}, can also be viewed as a special case of VRM.
Both methods aim to enhance the model’s robustness by increasing the diversity of the training data.

\subsection{Details of Loss Function}
\label{sec:detial_loss}

In this section, we provide the detailed loss function \cref{eq:virm}.
The vicinal features were defined in \cref{eq:z_tilde}, and the main idea of VIRM is to generalize features with the same label.
VIRM introduce a model $q_{\phi}(\mathbf{\xi}|\mathbf{\mathbf{z}}_i)$ to approximate the distribution $p(\mathbf{\xi} |\mathbf{z}_i)$ for all domains, 
and minimizing the KL divergence between $q_{\phi}(\mathbf{\xi}|\mathbf{\mathbf{z}}_i)$ and $p(\mathbf{\xi}|\mathbf{\mathbf{z}}_i)$.
The optimization goal of the SDA module is defined as:
\begin{equation}
	\phi = \underset{\phi}{\arg\max} D_{KL}(q_{\phi}(\mathbf{\xi}|\mathbf{\mathbf{z}}_i) || p(\mathbf{\xi}|\mathbf{\mathbf{z}}_i)).
\end{equation}

Assuming an edge distribution $ p(\mathbf{\xi})\sim \mathcal{N}(0, \mathbf{I})$ and $q_{\phi}(\mathbf{\xi}|\mathbf{z}) \sim \mathcal{N}(0, \mathbf{\sigma^2})$, 
the mean will be set to 0 because we expect to learn an offset, not an augmented feature.
$p(\mathbf{z}|\mathbf{\xi})$ is reconstructor.
Therefore, the optimizes the following objective function:
\begin{equation}
  \begin{aligned}
         \mathcal{L}(\phi;\mathbf{z})  = - \text{KL}(q_{\phi}(\mathbf{\xi}|\mathbf{z}) || p(\mathbf{\xi})) + \mathbb{E}_{\mathbf{\xi} \sim q_{\phi}(\mathbf{\xi}|\mathbf{z}) }(\log{ p(\mathbf{z}|\mathbf{\xi}) }).
 \end{aligned}
 \label{eq:virm_optim}
\end{equation}

The first term of \cref{eq:virm_optim} can be calculated easily. 
The second term estimates the features $\mathbf{z}$ given $\mathbf{\xi}$, which is more challenging to compute. 
Drawing inspiration from the design of Variational Autoencoders (VAE)~\cite{kingma2013auto}, 
we introduce a reconstruction network for learning this term, rewriting the second term as $ p_{\phi}(\mathbf{z}|\mathbf{\xi})$.
Then, we remark all parameters as $\phi$, we obtain the loss function as follows:

\begin{equation}
	\label{eq:virm_loss_appd}
  \begin{aligned}
     \mathcal{L}(\phi;\mathbf{z}) = -\text{KL}(q_{\phi}(\mathbf{\xi}|\mathbf{z}) || p(\mathbf{\xi})) 
      + \mathbb{E}_{\mathbf{\xi} \sim q_{\phi}(\mathbf{\xi}|\mathbf{z}) }(\log{ p_{\phi}(\mathbf{z}|\mathbf{\xi}) }).
  \end{aligned}
\end{equation}

The second part of \cref{eq:virm_loss_appd} depends on the model, and we use MSE loss. 
Assuming the marginal distribution $p(\mathbf{\xi})$ follows a normal distribution $\mathcal{N}(0, \mathbf{I})$ and $q_{\phi}(\mathbf{\xi}|\mathbf{z})$ also follows a normal distribution $\mathcal{N}(0, \boldsymbol{\sigma}^2)$, 
setting the mean to zero because we aim to learn the offset relative to the original rather than the augmented feature.
Thus, the loss function of VIRM is given by:
\begin{equation}
  \begin{aligned}
\mathcal{L}_{\phi} = - \frac{1}{2}\sum_{i=0}^n (1 + \log(\boldsymbol{\sigma}^2) - \boldsymbol{\sigma}^2 ) + \frac{1}{2n}\sum_{i=1}^n (\mathbf{\hat{z}_i} - \mathbf{z_i})^2,
\end{aligned}
\end{equation}
where $\boldsymbol{\sigma}^2$ is estimated variance, and $\mathbf{\hat{z}_i}$ is reconstructed feature using $\mathbf{\xi}$.
We employ the reparameterization trick to facilitate the computation of the loss function while ensuring the gradient flow for effective backpropagation.

\subsection{More Experiments}
\label{exp:abl_st}

\textbf{Potential component:}
We combined  domain-shared SDA with various IRM implementations, including IRMv1, VREx, and IB-IRM, and evaluated the performance on the ColoredMNIST dataset. 
The results are shown in \Cref{tab:abl_com}. 
VREx achieved the largest average accuracy improvement across all domains, making it the chosen IRM constraint for VRIM.

\begin{table}[t]
  \renewcommand\arraystretch{1.2}
  \centering
  \caption{
    Average accuracy results on the ColoredMNIST dataset according to the DomainBed benchmark\cite{gulrajani2020search}.
    SDA is utilized to combine with various IRM implementations.
    }
  \adjustbox{max width=0.45\textwidth}{%
  \begin{tabular}{lcccc}
  \hline
  \textbf{Method}  & \textbf{$+$90\%}    & \textbf{$+$80\%}    & \textbf{$-$90\%}      & \textbf{Avg.} \\
  \hline
  IRMv1 \cite{arjovsky2019invariant}   & 50.0   & 50.1  & 10.0   & 36.7  \\
  \quad + SDA    & 58.3        & 49.9        & 10.0        & \quad 39.4 (+2.7) \\
  \hline
  IB-IRM \cite{ahuja2021invariance}      & 56.9        & 49.8        & 10.1  & 38.9  \\
  \quad +  SDA    & 63.0       & 49.8       & 10.1        & \quad 41.0 (+2.1)  \\
  \hline
  VREx \cite{krueger2021out}            & 50.2        & 50.5        & 10.1        & 36.9   \\
  \quad +  SDA    & 50.6        & 59.7        & 10.1        &  \quad 40.1 (\red{+3.2}) \\
  \hline
\end{tabular}}
\label{tab:abl_com}
\end{table}

\begin{figure}[t]
  \centering
  \caption{
    Visualization of feature distributions using UMAP on the VLCS dataset (training set: VLS, test set: C).
  }
  \begin{subfigure}[b]{0.23\textwidth}
      \centering
      \includegraphics[width=\textwidth]{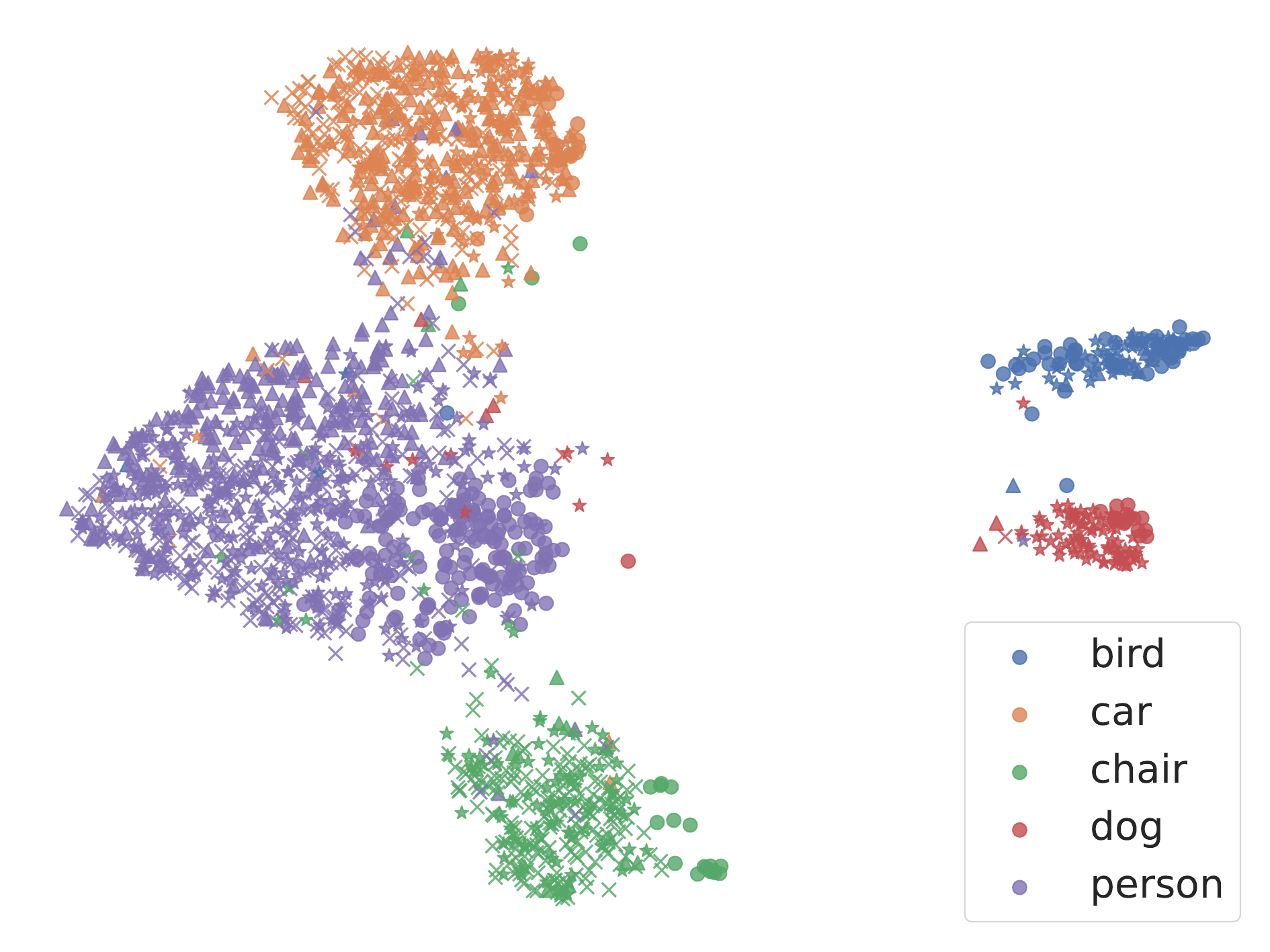}
      \caption{Original features}
      \label{fig: umap_original_features_filtered}
  \end{subfigure}
  \hfill
  \begin{subfigure}[b]{0.23\textwidth}
      \centering
      \includegraphics[width=\textwidth]{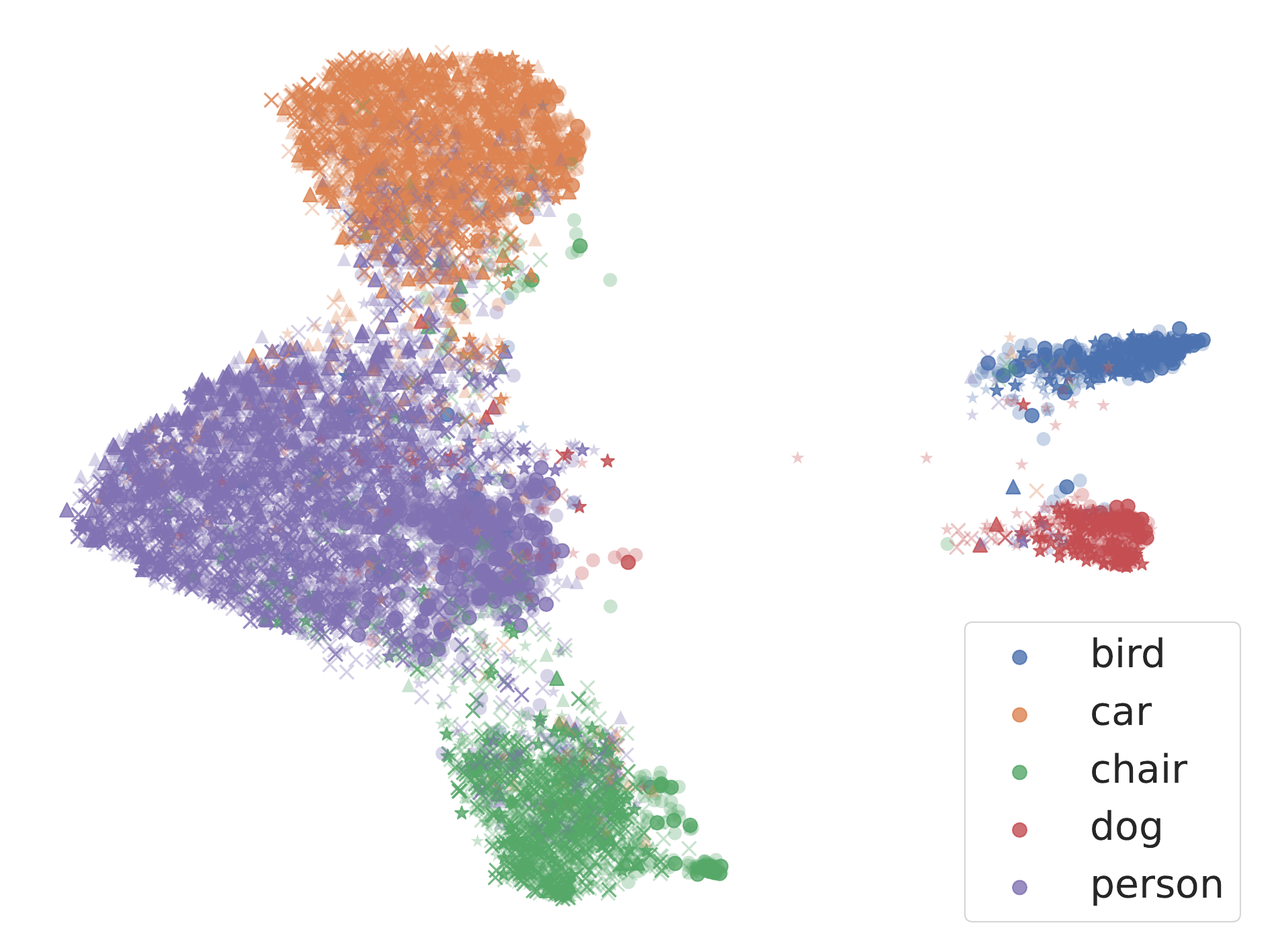}
      \caption{Augmented features}
      \label{fig: umap_augmented_features}
  \end{subfigure}
  \label{fig: vis_umap}
\end{figure}

\textbf{Visualization:}
We used UMAP~\cite{mcinnes2018umap} to visualize feature distributions across categories and domains.
\Cref{fig: umap_original_features_filtered} shows the original feature distribution, while \Cref{fig: umap_augmented_features} highlights the denser distribution of augmented features, encapsulating the original features.

\subsection{Details of Generalization Error Bound}
\label{sec: appd_bsda_rc}

\subsubsection{Rademacher Complexity}

The Rademacher complexity is a measure of the complexity of a function class, which can be used to derive generalization bounds for learning algorithms.
The Rademacher complexity of a function class $\mathcal{F}$ is defined as:
\begin{equation}
	\begin{aligned}
    R(\mathcal{F} ) = \mathbb{E}_{\sigma} \left[ \sup_{f \in \mathcal{F}} \frac{1}{m} \sum_{i=1}^m \sigma_i f(\mathbf{x}_i) \right],
	\end{aligned}
\end{equation}
where $\sigma_i$ are Rademacher variables, $\mathbf{x}_i$ are samples from the data distribution, and $m$ is the number of samples.

\subsubsection{Proof of \Cref{lemma:bound}}
\begin{lemma_supp}
	\label{supp_lamma:bound}
	Let $S$ be training set, hypothesis class $\mathbb{H}$, loss function $\ell$.
	Then:
	\begin{equation}
		\frac{1}{2} \mathbb{E}_S R(\bar{\mathcal{F}} \circ \mathcal{S}) 
		\le \mathbb{E}_S \sup_{f \in \mathcal{F}} |\mathbb{E} f - \hat{\mathbb{E}}f | 
		\le 2 \mathbb{E}_S R(\mathcal{F} \circ \mathcal{S}),
    \label{eq:bound}
	\end{equation}
	where $\hat{\mathbb{E}}f$ is the empirical expectation of $f$, and to simplify the notation, let $\mathcal{F}\overset{\mathrm{def}}{\operatorname*{=}}\ell\circ\mathcal{H}\overset{\mathrm{def}}{\operatorname*{=}}\{z\mapsto\ell(h,z):h\in\mathcal{H}\}$, $\bar{\mathcal{F}}=\{f-\mathbb{E}[f]\mid f\in\mathcal{F}\}$.
\end{lemma_supp}

\Cref{lemma:bound} establishes bounds on the supremum of the difference between the expected and empirical risks for a hypothesis class $\mathcal{H}$ with respect to a loss function $\ell$.
To prove the lemma, we decompose the argument into two parts, corresponding to the lower and upper bounds.
For convenience, we define the following notation:
$$
\Phi (S) = \sup_{f \in \mathcal{F}} |\mathbb{E} f - \hat{\mathbb{E}} f | ,
$$
where $S = (z_1, ..., z_m)$ a fixed sample of size $m$.

\begin{proof}[Lower Bound of \Cref{lemma:bound}]
  Let $S$ and $S^{\prime}$ be two samples differing by exactly one point, say $z_i$ in $S$ and $z^{\prime}_i$ in $S^{\prime}$. We have:
\begin{equation}
  \begin{aligned}
    \frac{1}{2} \mathbb{E}_S R(\bar{\mathcal{F}} \circ \mathcal{S}) 
    &= \frac{1}{2} \mathbb{E}_S \left[ \sup_{f \in \bar{\mathcal{F}} \circ \mathcal{S}} \frac{1}{m} \sum_{i=1}^m \sigma_i f(z_i) \right] \\
    &= \underset{\sigma,S,S^{\prime}}{\operatorname*{\mathbb{E}}}\left[\frac{1}{2} \underset{f \in \mathcal{F}}{\operatorname*{\sup}}\frac1m\sum_{i=1}^m\sigma_i(f(z_i)- \mathbb{E}f(z_i^{\prime}))\right] \\
    &\leq \underset{\sigma,S,S^{\prime}}{\operatorname*{\mathbb{E}}}\left[\frac{1}{2} \underset{f \in \mathcal{F}}{\operatorname*{\sup}}\frac1m\sum_{i=1}^m\sigma_i(f(z_i)- f(z_i^{\prime}))\right]. \\
  \end{aligned}
\end{equation}
The above derivation assumes $\mathbb{E}f = 0$, simplifying $\mathbb{E}f(z_i^{\prime})$ to zero, and enabling the use of sample differences in the third line.
Then by triangular inequality, we have:
  \begin{equation}
    \begin{aligned}
      & \frac{1}{2} \underset{f \in \mathcal{F}}{\operatorname*{\sup}}\frac1m\sum_{i=1}^m\ (f(z_i)- f(z_i^{\prime})) \\
      &\leq \frac{1}{2} \underset{f \in \mathcal{F}}{\operatorname*{\sup}}\frac1m\sum_{i=1}^m\ (f(z_i)- \mathbb{E}f(z)) + \frac{1}{2} \underset{f \in \mathcal{F}}{\operatorname*{\sup}}\frac1m\sum_{i=1}^m\ (f(z_i^{\prime})- \mathbb{E}f(z)). \\
    \end{aligned}
  \end{equation}
\end{proof}

Taking expectation on both sides and noticing that $z_i$ and $z^{\prime}_i$ are identically distributed give the lower bound in \Cref{eq:bound}.

\begin{proof}[Upper Bound of \Cref{lemma:bound}]
  We have:
\begin{equation}
  \begin{aligned}
    \operatorname{\mathbb{E}_S}[\Phi(S)]& =\operatorname{\mathbb{E}}_S\left[\sup_{f \in \mathcal{F}}\operatorname{\mathbb{E}}[f]-\operatorname{\hat{\mathbb{E}}}_S(f)\right] \\
      &=\operatorname{\mathbb{E}}_S\left[\sup_{f \in \mathcal{F}}\operatorname{\mathbb{E}}\left[\operatorname{\widehat{\mathbb{E}}}_{S^{\prime}}(f)-\operatorname{\widehat{\mathbb{E}}}_S(f)\right]\right] \\
      &\leq\underset{S,S^{\prime}}{\operatorname*{\mathbb{E}}}\left[\underset{f \in \mathcal{F}}{\operatorname*{\sup}}\widehat{\mathbb{\mathbb{E}}}_{S^{\prime}}(f)-\widehat{\mathbb{\mathbb{E}}}_S(f)\right] \\
      &=\underset{S,S^{\prime}}{\operatorname*{\mathbb{E}}}\left[\underset{f \in \mathcal{F}}{\operatorname*{\sup}}\frac1m\sum_{i=1}^m(f(z_i^{\prime})-f(z_i))\right] \\
      &=\underset{\sigma,S,S^{\prime}}{\operatorname*{\mathbb{E}}}\left[\underset{f \in \mathcal{F}}{\operatorname*{\sup}}\frac1m\sum_{i=1}^m\sigma_i(f(z_i^{\prime})-f(z_i))\right] \\
      &\leq\underset{\sigma,S^{\prime}}{\operatorname*{\mathbb{E}}}\left[\underset{f \in \mathcal{F}}{\operatorname*{\sup}}\frac1m\sum_{i=1}^m\sigma_if(z_i^{\prime})\right] +  \underset{\sigma,S}{\operatorname*{\mathbb{E}}}\left[\underset{f \in \mathcal{F}}{\operatorname*{\sup}}\frac1m\sum_{i=1}^m-\sigma_if(z_i)\right] \\
      &=2\operatorname{\mathbb{E}}_{\boldsymbol{\sigma},S}\Big[\sup_{f \in \mathcal{F}}\frac1m\sum_{i=1}^m\sigma_if(z_i)\Big] \\ 
      &= 2 \mathbb{E}_S R(\mathcal{F} \circ \mathcal{S}).
  \end{aligned}
\end{equation}
\end{proof}

\subsubsection{Proof of \Cref{lemma:rc}}
\begin{lemma_supp}
  \label{supp_lamma:rc}
  Let $S = (\mathbf{x}_1,...,\mathbf{x}_m)$ be vectors in a Hilbert space. Define: $\mathcal{H} \circ S = \{ (\langle\mathbf{w},\mathbf{x}_1\rangle, ...,\langle\mathbf{w},\mathbf{x}_m\rangle):||\mathbf{w}||_2 \le 1 \}$. 
  Then:
	\begin{equation}
			R(\mathcal{H}\circ S)~\leq~ \sqrt{\frac{\max_i\|\mathbf{x}_i\|_2^2}{m}}.
      \label{eq:rc_baseline}
	\end{equation}
\end{lemma_supp}

\begin{proof}
    Using Cauchy-Schwarz inequality we know that for any two vectors $\mathbf{w}$ and $\mathbf{v}$, $\langle \mathbf{w}, \mathbf{v} \rangle \le \|\mathbf{w}\| \|\mathbf{v}\|$. Therefore:
    \begin{equation}
        \begin{aligned}
            mR(\mathcal{H} \circ S) 
            &= \mathbb{E}_{\boldsymbol{\sigma}} \left[ \sup_{\mathbf{a} \in \mathcal{H} \circ S} \sum_{i=1}^m \sigma_i a_i \right] \\
            &= \mathbb{E} \left[ \sup_{\mathbf{w} : \|\mathbf{w}\| \leq 1} \sum_{i=1}^m \sigma_i \langle \mathbf{w}, \mathbf{x}_i \rangle \right] \\
            &= \mathbb{E}_{\boldsymbol{\sigma}} \left[ \sup_{\mathbf{w} : \|\mathbf{w}\| \leq 1} \langle \mathbf{w}, \sum_{i=1}^m \sigma_i \mathbf{x}_i \rangle \right] \\
            &\leq \mathbb{E}_{\boldsymbol{\sigma}} \left[ \left\|\sum_{i=1}^m \sigma_i \mathbf{x}_i \right\|_2 \right].
        \end{aligned}
    \end{equation}

    Next, by Jensen's inequality we have that:
    \[
    \begin{aligned}
        \mathbb{E}_{\boldsymbol{\sigma}} \left[ \left\|\sum_{i=1}^m \sigma_i \mathbf{x}_i \right\|_2 \right] &= \mathbb{E}_{\boldsymbol{\sigma}} \left[ \left( \left\|\sum_{i=1}^m \sigma_i \mathbf{x}_i \right\|_2^2 \right)^{1/2} \right] \\
        &\leq \left( \mathbb{E}_{\boldsymbol{\sigma}} \left[ \left\|\sum_{i=1}^m \sigma_i \mathbf{x}_i \right\|_2^2 \right] \right)^{1/2}.
    \end{aligned}
    \]
    
    Finally, since $\sigma_1, ..., \sigma_m$ are independent, we have:
    \[
    \begin{aligned}
        \mathbb{E}_{\boldsymbol{\sigma}} \left[ \left\|\sum_{i=1}^m \sigma_i \mathbf{x}_i \right\|_2^2 \right] &= \mathbb{E}_{\boldsymbol{\sigma}} \left[ \sum_{i,j} \sigma_i \sigma_j \langle \mathbf{x}_i, \mathbf{x}_j \rangle \right] \\
        &= \sum_{i \neq j} \langle \mathbf{x}_i, \mathbf{x}_j \rangle \underset{\boldsymbol{\sigma}}{\operatorname*{\mathbb{E}}} \left[\sigma_i \sigma_j\right] + \sum_{i=1}^m \langle \mathbf{x}_i, \mathbf{x}_i \rangle \underset{\boldsymbol{\sigma}}{\operatorname*{\mathbb{E}}} \left[\sigma_i^2\right] \\
        &= \sum_{i=1}^m \left\| \mathbf{x}_i \right\|_2^2 \\
        &\leq m \max_i \|\mathbf{x}_i\|_2^2.
    \end{aligned}
    \]
\end{proof}

\subsubsection{Proof of \Cref{theorem:bsda_rc}}
\begin{theorem_supp}
    Let $\mathcal{H}$ be the hypothesis class, $S$ be the training set, and $\mathcal{H}\circ S$ be the linear class.
    Then the Rademacher complexity of the \Cref{eq:z_tilde} is bounded as follows:
    \begin{equation}    
                \tilde{R}(\mathcal{H}\circ S) ~\leq~ \sqrt{ \frac{(\max_i \left\| \mathbf{x}_i  \right\|_2^2 + k)}{n} } \\
    \end{equation}
    where $n$ is the number of samples augmentation, is a multiple of $m$ generally, $k$ is feature dimension.
\end{theorem_supp}

\begin{proof}
By the Cauchy-Schwarz inequality, for any two vectors \(\mathbf{w}\) and \(\mathbf{v}\), we have \(\langle \mathbf{w}, \mathbf{v} \rangle \le \|\mathbf{w}\| \|\mathbf{v}\|\). Therefore:
\[
\begin{aligned}
n\tilde{R}(\mathcal{H}\circ S)  &= \mathbb{E}_{\boldsymbol{\sigma}} \left[ \sup_{\mathbf{a} \in \mathcal{H} \circ S} \sum_{i=1}^n \sigma_i a_i \right] \\
&= \mathbb{E} \left[ \sup_{\mathbf{w} : \|\mathbf{w}\| \leq 1} \sum_{i=1}^n \sigma_i \langle \mathbf{w}, \Phi(\mathbf{x}_i) \rangle \right] \\
&= \mathbb{E}_{\boldsymbol{\sigma}} \left[ \sup_{\mathbf{w} : \|\mathbf{w}\| \leq 1} \langle \mathbf{w}, \sum_{i=1}^n \sigma_i \Phi(\mathbf{x}_i) \rangle \right] \\
&\leq \mathbb{E}_{\boldsymbol{\sigma}} \left[ \left\|\sum_{i=1}^n \sigma_i \Phi(\mathbf{x}_i) \right\|_2 \right].
\end{aligned}
\]

Now, by Jensen's inequality, we have:
\[
\begin{aligned}
\mathbb{E}_{\boldsymbol{\sigma}} \left[ \left\|\sum_{i=1}^n \sigma_i \Phi(\mathbf{x}_i) \right\|_2 \right] &= \mathbb{E}_{\boldsymbol{\sigma}} \left[ \left( \left\|\sum_{i=1}^n \sigma_i \Phi(\mathbf{x}_i) \right\|_2^2 \right)^{1/2} \right] \\
&\leq \left( \mathbb{E}_{\boldsymbol{\sigma}} \left[ \left\|\sum_{i=1}^n \sigma_i \Phi(\mathbf{x}_i) \right\|_2^2 \right] \right)^{1/2}.
\end{aligned}
\]

Since \(\sigma_1, ..., \sigma_n\) are independent, we have:
\[
\begin{aligned}
&\mathbb{E}_{\boldsymbol{\sigma}} \left[ \left\|\sum_{i=1}^n \sigma_i \Phi(\mathbf{x}_i) \right\|_2^2 \right] \\ 
&= \mathbb{E}_{\boldsymbol{\sigma}} \left[ \sum_{i,j} \sigma_i \sigma_j \langle \Phi(\mathbf{x}_i), \Phi(\mathbf{x}_j) \rangle \right] \\
&= \sum_{i \neq j} \langle \Phi(\mathbf{x}_i), \Phi(\mathbf{x}_j) \rangle \underset{\boldsymbol{\sigma}}{\operatorname*{\mathbb{E}}} \left[\sigma_i \sigma_j\right] + \sum_{i=1}^n \langle \Phi(\mathbf{x}_i), \Phi(\mathbf{x}_i) \rangle \underset{\boldsymbol{\sigma}}{\operatorname*{\mathbb{E}}} \left[\sigma_i^2\right] \\
&= \sum_{i=1}^n \left\| \Phi(\mathbf{x}_i) \right\|_2^2.
\end{aligned}
\]

Let \(\Phi(\mathbf{x}) = \mathbf{x} + \mathbf{\xi}\), where \(p(\mathbf{\xi}) = \mathcal{N}(0, I)\) and \(p(\mathbf{\xi} | \mathbf{x}) = \mathcal{N}(0, \boldsymbol{\sigma}^2)\). 
\(\mathbf{\xi}\) is sampled from the distribution \(q_{\phi}(\mathbf{\xi} | \mathbf{x})\). 
Then, the above equation becomes:
\begin{equation}
  \begin{aligned}
    \sum_{i=1}^n \left\| \Phi(\mathbf{x}_i) \right\|_2^2 &= \sum_{i=1}^n \left\| \mathbf{x}_i + \mathbf{\xi}_i \right\|_2^2 \\
    &= \sum_{i=1}^n \left[\|\mathbf{x}_i\|_2^2 + 2 \mathbf{x}_i^T \mathbf{\xi}_i + \|\mathbf{\xi}_i\|_2^2\right] \\
    \end{aligned}
    \label{eq:rc_value}
\end{equation}


In \Cref{eq:rc_value}, the distribution of $\mathbf{\xi}$ can be analyzed from both a local and a global perspective. Locally, $\mathbf{\xi}$ is sampled conditioned on each data point $\mathbf{x}$, following $p(\mathbf{\xi} | \mathbf{x}) = \mathcal{N}(0, \boldsymbol{\sigma}^2 \mathbf{I})$, where $|\mathbf{\xi}|_2^2$ has an expected value of $k \cdot \boldsymbol{\sigma}^2$, and $k$ represents the dimension of the feature space. 
This reflects the noise characteristics specific to each data instance.

Globally, however, $\mathbf{\xi}$ is drawn from the marginal distribution $p(\mathbf{\xi}) = \mathcal{N}(0, \mathbf{I})$, which is independent of $\mathbf{x}$. Here, $|\mathbf{\xi}|_2^2$ has an expected value of $k$, since the global variance of $\mathbf{\xi}$ is normalized to 1. 
This global perspective captures the aggregate error distribution over all data points, providing an upper bound for the local variability. 
By focusing on the marginal distribution $p(\mathbf{\xi})$, we ensure that our analysis considers the variance uniformly across the entire dataset, which is critical for deriving robust generalization bounds.
Thus:
\begin{equation}
\begin{aligned}
  \tilde{R}(\mathcal{H}\circ S)  &= \frac{1}{n} \sqrt{\sum_{i=1}^n \left[\|\mathbf{x}_i\|_2^2 + k\right]} \\
&= \frac{1}{n} \sqrt{n (\max_i \|\mathbf{x}_i\|_2^2 + k)} \\
&= \sqrt{\frac{\max_i \|\mathbf{x}_i\|_2^2 + k}{n}}.
\label{eq:rc_virm}
\end{aligned}
\end{equation}
\end{proof}

In our analysis, we compare two expressions for the Rademacher complexity bound. 
The first expression, as shown in \Cref{eq:rc_baseline}, considers the Rademacher complexity in the context of augmented data, where the sample size \(n\) is significantly larger than both the number of training domains \(m\) and the noise dimension \(k\). 
This results in a complexity bound of the form:

\[
  \tilde{R}(\mathcal{H}\circ S) = \sqrt{\frac{\max_i \|\mathbf{x}_i\|_2^2 + k}{n}}.
\]

Since \(n > m\) and \(n \gg k\), the impact of \(k\) becomes negligible, and the bound can be approximated by:

\[
  \tilde{R}(\mathcal{H}\circ S) \approx \sqrt{\frac{\max_i \|\mathbf{x}_i\|_2^2}{n}}.
\]

On the other hand, the second expression, given in \Cref{eq:rc_virm}, provides an upper bound on the Rademacher complexity:

\[
  \tilde{R}(\mathcal{H}\circ S) \leq \sqrt{\frac{\max_i \|\mathbf{x}_i\|_2^2}{m}}.
\]

Given that \(n > m\), the complexity bound in the first expression is typically smaller than or equal to the bound in the second expression. 
Therefore, the second expression serves as an upper bound, while the first expression provides a more refined estimate of the Rademacher complexity when considering augmented data. This analysis suggests that the larger sample size in the augmented dataset helps in controlling the complexity, thereby improving generalization.

\subsubsection{Analysis}

In our analysis, we compare the Rademacher complexity bounds for the baseline method and our method  with respect to their impact on generalization error. 
The formula

\[
\frac{1}{2} \mathbb{E}_S R(\bar{\mathcal{F}} \circ \mathcal{S}) 
\le \mathbb{E}_S \sup_{f \in \mathcal{F}} |\mathbb{E} f - \hat{\mathbb{E}}f | 
\le 2 \mathbb{E}_S R(\mathcal{F} \circ \mathcal{S})
\]

provides a general framework for bounding the generalization error in terms of Rademacher complexity. 
The left-hand side gives a lower bound based on the complexity of the model class \(\bar{\mathcal{F}}\), and the right-hand side provides an upper bound based on \(\mathcal{F}\). 
For the baseline method, the complexity bound is inversely proportional to the number of training domains \(m\), leading to a relatively larger upper bound. 
In contrast, our method leverages a larger augmented sample size \(n\), resulting in a smaller Rademacher complexity bound, as \(n \gg m\). 
This allows our method to achieve a tighter upper bound, thus reducing the generalization error. 
Additionally, the impact of the noise term \(k\) in our method is minimal compared to the effect of \(n\), further enhancing the generalization capacity of our approach. 
Consequently, our method provides both a smaller lower bound and a tighter upper bound on the generalization error, demonstrating its superior capacity for generalization compared to the baseline method.

\subsection{Full DomainBed results}
\label{sec:full_result}

In this section, we provide the full results of the DomainBed benchmark~\cite{gulrajani2020search} on the PACS~\cite{pacs}, VLCS~\cite{vlcs}, OfficeHome~\cite{officehome} and TerraIncognita~\cite{beery2018recognition} datasets.
The results are shown in \cref{tab:pacs}, \cref{tab:vlcs}, \cref{tab:oh} and \cref{tab:tr}, respectively.
The results are obtained using the default hyper-parameter settings in DomainBed with leave-one-domain-out cross-validation.
The proposed method outperforms the state-of-the-art methods on both datasets, demonstrating the effectiveness of the proposed method.

\begin{table*}[htbp]
    \centering
    \renewcommand\arraystretch{1.1}
    \caption{
      Average accuracy results on the PACS~\cite{pacs} dataset according to the DomainBed benchmark~\cite{gulrajani2020search}.
      }
    \adjustbox{max width=\textwidth}{%
    \begin{tabular}{lcccccc}
        \hline
        \textbf{Method} & \textbf{Art} & \textbf{Cartoon} & \textbf{Photo} & \textbf{Sketch} & \textbf{Avg.} \\
        \hline
        HYPO & 74.7  & 77.7 &  93.9 & 69.2 &  78.9 \\
        DecAug & 79.0 & 79.6 & 75.6 & 95.3 & 82.4 \\
        VNE & 77.6 $\pm$  0.8 & 76.2 $\pm$ 0.9 & 94.8 $\pm$  0.5 & 73.5 $\pm$ 1.2 & 80.5 \\
        ERM  & 78.0 $\pm$ 1.3 & 73.4 $\pm$ 0.8 & 94.1 $\pm$ 0.4 & 73.6 $\pm$ 2.2 & 79.8  \\
        IRM  & 76.9 $\pm$ 2.6 & 75.1 $\pm$ 0.7 & 94.3 $\pm$ 0.4 & 77.4 $\pm$ 0.4 & 80.9 \\
        GroupGRO  & 77.7 $\pm$ 2.6 & 76.4 $\pm$ 0.3 & 94.0 $\pm$ 0.3 & 74.8 $\pm$ 1.3 & 80.7 \\
        Mixup  & 79.3 $\pm$ 1.1 & 74.2 $\pm$ 2.0 & 94.9 $\pm$ 0.3 & 68.3 $\pm$ 2.7 & 79.2  \\
        MLDG  & 78.4 $\pm$ 0.7 & 75.1 $\pm$ 0.5 & 94.8 $\pm$ 0.4 & 76.7 $\pm$ 0.8 & 81.3  \\
        CORAL  & 81.5 $\pm$ 0.5 & 75.4 $\pm$ 0.7 & 95.2 $\pm$ 0.5 & 74.8 $\pm$ 0.4 & 81.7 \\
        MMD  & 81.3 $\pm$ 0.6 & 75.5 $\pm$ 1.0 & 94.0 $\pm$ 0.5 & 74.3 $\pm$ 1.5 & 81.3 \\
        DANN  & 79.0 $\pm$ 0.6 & 72.5 $\pm$ 0.7 & 94.4 $\pm$ 0.5 & 70.8 $\pm$ 3.0 & 79.2 \\
        CDANN  & 80.4 $\pm$ 0.8 & 73.7 $\pm$ 0.3 & 93.1 $\pm$ 0.6 & 74.2 $\pm$ 1.7 & 80.3 \\
        MTL  & 78.7 $\pm$ 0.6 & 73.4 $\pm$ 1.0 & 94.1 $\pm$ 0.6 & 74.4 $\pm$ 3.0 & 80.1 \\
        SagNet  & 82.9 $\pm$ 0.4 & 73.2 $\pm$ 1.1 & 94.6 $\pm$ 0.5 & 76.1 $\pm$ 1.8 & 81.7 \\
        ARM  & 79.4 $\pm$ 0.6 & 75.0 $\pm$ 0.7 & 94.3 $\pm$ 0.6 & 73.8 $\pm$ 0.6 & 80.6 \\
        VREx  & 74.4 $\pm$ 0.7 & 75.0 $\pm$ 0.4 & 93.3 $\pm$ 0.3 & 78.1 $\pm$ 0.9 & 80.2 \\
        RSC  & 78.5 $\pm$ 1.1 & 73.3 $\pm$ 0.9 & 93.6 $\pm$ 0.6 & 76.5 $\pm$ 1.4 & 80.5 \\
        SelfReg  & 82.5 $\pm$ 0.8 & 74.4 $\pm$ 1.5 & 95.4 $\pm$ 0.5 & 74.9 $\pm$ 1.3 & 81.8 \\
        MixStyle  & 82.6 $\pm$ 1.2 & 76.3 $\pm$ 0.4 & 94.2 $\pm$ 0.3 & 77.5 $\pm$ 1.3 & 82.6 \\
        Fish  & 80.9 $\pm$ 1.0 & 75.9 $\pm$ 0.4 & 95.0 $\pm$ 0.4 & 76.2 $\pm$ 1.0 & 82.0 \\
        SD  & 83.2 $\pm$ 0.6 & 74.6 $\pm$ 0.3 & 94.6 $\pm$ 0.1 & 75.1 $\pm$ 1.6 & 81.9 \\
        CAD  & 83.9 $\pm$ 0.8 & 74.2 $\pm$ 0.4 & 94.6 $\pm$ 0.4 & 75.0 $\pm$ 1.2 & 81.9 \\
        CondCAD  & 79.7 $\pm$ 1.0 & 74.2 $\pm$ 0.9 & 94.6 $\pm$ 0.4 & 74.8 $\pm$ 1.4 & 80.8 \\
        Fishr  & 81.2 $\pm$ 0.4 & 75.8 $\pm$ 0.8 & 94.3 $\pm$ 0.3 & 73.8 $\pm$ 0.6 & 81.3 \\
        MIRO  & 79.3 $\pm$ 0.6 & 68.1 $\pm$ 2.5 & 95.5 $\pm$ 0.3 & 60.6 $\pm$ 3.1 & 75.9 \\
        DGRI & 82.4 $\pm$ 1.0 & 76.7 $\pm$ 0.6 & 95.3 $\pm$ 0.1 & 76.7 $\pm$ 0.3 & 82.8 \\
        W2D & 81.2 $\pm$ 0.8 & 72.8 $\pm$ 1.8 & 93.9 $\pm$ 0.2 & 75.9 $\pm$ 0.2 & 80.9 \\
        \hline
        Ours & 83.3 $\pm$ 1.6 & 76.3 $\pm$ 0.9 & 95.3 $\pm$ 0.4 & 75.7 $\pm$ 0.2 & 82.6 \\
        \hline
      \end{tabular}
  }
  \label{tab:pacs}
  \end{table*}

\begin{table*}[htbp]
    \centering
    \renewcommand\arraystretch{1.1}
    \caption{
        Average accuracy results on the VLCS~\cite{vlcs} dataset according to the DomainBed benchmark~\cite{gulrajani2020search}.
      }
    \adjustbox{max width=\textwidth}{%
    \begin{tabular}{lccccc}
        \hline
        \textbf{Method} & \textbf{Caltech} & \textbf{LabelMe} & \textbf{Sun} & \textbf{VOC} & \textbf{Avg.} \\
        \hline
        HYPO & 98.4 & 61.5 & 71.2 & 73.0 & 76.0 \\
        DecAug & 95.6 & 62.4 & 68.1 & 72.3 & 74.6 \\
        VNE & 97.8 $\pm$ 0.4 & 63.7 $\pm$ 0.4 & 70.8 $\pm$ 0.4 & 74.5 $\pm$  0.8 & 76.7\\
        ERM  & 97.7 $\pm$ 0.3 & 62.1 $\pm$ 0.9 & 70.3 $\pm$ 0.9 & 73.2 $\pm$ 0.7 & 75.8 \\
        IRM  & 96.1 $\pm$ 0.8 & 62.5 $\pm$ 0.3 & 69.9 $\pm$ 0.7 & 72.0 $\pm$ 1.4 & 75.1 \\
        GroupGRO  & 96.7 $\pm$ 0.6 & 61.7 $\pm$ 1.5 & 70.2 $\pm$ 1.8 & 72.9 $\pm$ 0.6 & 75.4 \\
        Mixup  & 95.6 $\pm$ 1.5 & 62.7 $\pm$ 0.4 & 71.3 $\pm$ 0.8 & 75.4 $\pm$ 0.2 & 76.2 \\
        MLDG  & 95.8 $\pm$ 0.5 & 63.3 $\pm$ 0.8 & 68.5 $\pm$ 0.5 & 73.1 $\pm$ 0.8 & 75.2 \\
        CORAL  & 96.5 $\pm$ 0.3 & 62.8 $\pm$ 0.1 & 69.1 $\pm$ 0.6 & 73.8 $\pm$ 1.0 & 75.5 \\
        MMD  & 96.0 $\pm$ 0.8 & 64.3 $\pm$ 0.6 & 68.5 $\pm$ 0.6 & 70.8 $\pm$ 0.1 & 74.9 \\
        DANN  & 97.2 $\pm$ 0.1 & 63.3 $\pm$ 0.6 & 70.2 $\pm$ 0.9 & 74.4 $\pm$ 0.2 & 76.3 \\
        CDANN  & 95.4 $\pm$ 1.2 & 62.6 $\pm$ 0.6 & 69.9 $\pm$ 1.3 & 76.2 $\pm$ 0.5 & 76.0 \\
        MTL  & 94.4 $\pm$ 2.3 & 65.0 $\pm$ 0.6 & 69.6 $\pm$ 0.6 & 71.7 $\pm$ 1.3 & 75.2 \\
        SagNet  & 94.9 $\pm$ 0.7 & 61.9 $\pm$ 0.7 & 69.6 $\pm$ 1.3 & 75.2 $\pm$ 0.6 & 75.4 \\
        ARM  & 96.9 $\pm$ 0.5 & 61.9 $\pm$ 0.4 & 71.6 $\pm$ 0.1 & 73.3 $\pm$ 0.4 & 75.9 \\
        VREx  & 96.2 $\pm$ 0.0 & 62.5 $\pm$ 1.3 & 69.3 $\pm$ 0.9 & 73.1 $\pm$ 1.2 & 75.3 \\
        RSC  & 96.2 $\pm$ 0.0 & 63.6 $\pm$ 1.3 & 69.8 $\pm$ 1.0 & 72.0 $\pm$ 0.4 & 75.4 \\
        SelfReg  & 95.8 $\pm$ 0.6 & 63.4 $\pm$ 1.1 & 71.1 $\pm$ 0.6 & 75.3 $\pm$ 0.6 & 76.4 \\
        MixStyle  & 97.3 $\pm$ 0.3 & 61.6 $\pm$ 0.1 & 70.4 $\pm$ 0.7 & 71.3 $\pm$ 1.9 & 75.2 \\
        Fish  & 97.4 $\pm$ 0.2 & 63.4 $\pm$ 0.1 & 71.5 $\pm$ 0.4 & 75.2 $\pm$ 0.7 & 76.9 \\
        SD  & 96.5 $\pm$ 0.4 & 62.2 $\pm$ 0.0 & 69.7 $\pm$ 0.9 & 73.6 $\pm$ 0.4 & 75.5 \\
        CAD  & 94.5 $\pm$ 0.9 & 63.5 $\pm$ 0.6 & 70.4 $\pm$ 1.2 & 72.4 $\pm$ 1.3 & 75.2 \\
        CondCAD  & 96.5 $\pm$ 0.8 & 62.6 $\pm$ 0.4 & 69.1 $\pm$ 0.2 & 76.0 $\pm$ 0.2 & 76.1 \\
        Fishr  & 97.2 $\pm$ 0.6 & 63.3 $\pm$ 0.7 & 70.4 $\pm$ 0.6 & 74.0 $\pm$ 0.8 & 76.2 \\
        MIRO  & 97.5 $\pm$ 0.2 & 62.0 $\pm$ 0.5 & 71.3 $\pm$ 1.0 & 74.8 $\pm$ 0.6 & 76.4 \\
        DGRI & 96.7 $\pm$ 0.5 & 63.2 $\pm$ 1.0 & 70.3 $\pm$ 0.8 & 73.4 $\pm$ 0.3 & 75.9 \\
        W2D & 93.9 $\pm$ 0.7 & 64.7 $\pm$ 1.1 & 66.2 $\pm$ 0.6 & 72.9 $\pm$ 2.4 & 74.4 \\
        \hline
        Ours & 96.8 $\pm$ 0.2 & 64.9 $\pm$ 0.9 & 72.1 $\pm$ 0.2 & 76.3 $\pm$ 0.2 & 77.5 \\
        \hline
      \end{tabular}
  }
  \label{tab:vlcs}
  \end{table*}

\begin{table*}[htbp]
  \renewcommand\arraystretch{1.1}
    \centering
    \caption{
      Average accuracy results on the OfficeHome~\cite{officehome} dataset according to the DomainBed benchmark~\cite{gulrajani2020search}.
      }
    \adjustbox{max width=\textwidth}{%
    \begin{tabular}{lccccc}
        \hline
        \textbf{Method} & \textbf{Art} & \textbf{Clipart} & \textbf{Product} & \textbf{Real} & \textbf{Avg.} \\
        \hline
        HYPO & 56.5 & 46.5 & 72.1 &  73.9 & 62.2 \\
        DecAug & 56.8 & 48.2 & 72.3 & 73.8 & 62.8 \\
        VNE & 50.6 $\pm$  0.8 & 45.2 $\pm$  0.1 & 67.4 $\pm$  0.2 & 69.2 $\pm$  0.2 & 58.1 \\
        ERM  & 52.2 $\pm$ 0.2 & 48.7 $\pm$ 0.5 & 69.9 $\pm$ 0.5 & 71.7 $\pm$ 0.5 & 60.6 \\
        IRM  & 49.7 $\pm$ 0.2 & 46.8 $\pm$ 0.5 & 67.5 $\pm$ 0.4 & 68.1 $\pm$ 0.6 & 58.0 \\
        GroupGRO  & 52.6 $\pm$ 1.1 & 48.2 $\pm$ 0.9 & 69.9 $\pm$ 0.4 & 71.5 $\pm$ 0.8 & 60.6 \\
        Mixup  & 54.0 $\pm$ 0.7 & 49.3 $\pm$ 0.3 & 70.7 $\pm$ 0.7 & 72.6 $\pm$ 0.3 & 61.7 \\
        MLDG  & 53.1 $\pm$ 0.3 & 48.4 $\pm$ 0.3 & 70.5 $\pm$ 0.7 & 71.7 $\pm$ 0.4 & 60.9 \\
        CORAL  & 55.1 $\pm$ 0.7 & 49.7 $\pm$ 0.9 & 71.8 $\pm$ 0.2 & 73.1 $\pm$ 0.5 & 62.4 \\
        MMD  & 50.9 $\pm$ 1.0 & 48.7 $\pm$ 0.3 & 69.3 $\pm$ 0.7 & 70.7 $\pm$ 1.3 & 59.9 \\
        DANN  & 51.8 $\pm$ 0.5 & 47.1 $\pm$ 0.1 & 69.1 $\pm$ 0.7 & 70.2 $\pm$ 0.7 & 59.5 \\
        CDANN  & 51.4 $\pm$ 0.5 & 46.9 $\pm$ 0.6 & 68.4 $\pm$ 0.5 & 70.4 $\pm$ 0.4 & 59.3 \\
        MTL  & 51.6 $\pm$ 1.5 & 47.7 $\pm$ 0.5 & 69.1 $\pm$ 0.3 & 71.0 $\pm$ 0.6 & 59.9 \\
        SagNet  & 55.3 $\pm$ 0.4 & 49.6 $\pm$ 0.2 & 72.1 $\pm$ 0.4 & 73.2 $\pm$ 0.4 & 62.5 \\
        ARM  & 51.3 $\pm$ 0.9 & 48.5 $\pm$ 0.4 & 68.0 $\pm$ 0.3 & 70.6 $\pm$ 0.1 & 59.6 \\
        VREx  & 51.1 $\pm$ 0.3 & 47.4 $\pm$ 0.6 & 69.0 $\pm$ 0.4 & 70.5 $\pm$ 0.4 & 59.5 \\
        RSC  & 49.0 $\pm$ 0.1 & 46.2 $\pm$ 1.5 & 67.8 $\pm$ 0.7 & 70.6 $\pm$ 0.3 & 58.4 \\
        SelfReg  & 55.1 $\pm$ 0.8 & 49.2 $\pm$ 0.6 & 72.2 $\pm$ 0.3 & 73.0 $\pm$ 0.3 & 62.4 \\
        MixStyle  & 50.8 $\pm$ 0.6 & 51.4 $\pm$ 1.1 & 67.6 $\pm$ 1.3 & 68.8 $\pm$ 0.5 & 59.6 \\
        Fish  & 54.6 $\pm$ 1.0 & 49.6 $\pm$ 1.0 & 71.3 $\pm$ 0.6 & 72.4 $\pm$ 0.2 & 62.0 \\
        SD  & 55.0 $\pm$ 0.4 & 51.3 $\pm$ 0.5 & 72.5 $\pm$ 0.2 & 72.7 $\pm$ 0.3 & 62.9 \\
        CAD  & 52.1 $\pm$ 0.6 & 48.3 $\pm$ 0.5 & 69.7 $\pm$ 0.3 & 71.9 $\pm$ 0.4 & 60.5 \\
        CondCAD  & 53.3 $\pm$ 0.6 & 48.4 $\pm$ 0.2 & 69.8 $\pm$ 0.9 & 72.6 $\pm$ 0.1 & 61.0 \\
        Fishr  & 52.6 $\pm$ 0.9 & 48.6 $\pm$ 0.3 & 69.9 $\pm$ 0.6 & 72.4 $\pm$ 0.4 & 60.9 \\
        MIRO  & 57.4 $\pm$ 0.9 & 49.5 $\pm$ 0.3 & 74.0 $\pm$ 0.1 & 75.6 $\pm$ 0.2 & 64.1 \\
        DGRI & 56.6 $\pm$ 0.7 & 50.3 $\pm$ 0.6 & 72.5 $\pm$ 0.0 & 73.8 $\pm$ 0.3 & 63.3 \\
        W2D & 51.2 $\pm$ 1.0 & 47.0 $\pm$ 0.2 & 67.5 $\pm$ 0.3 & 69.4 $\pm$ 0.1 & 58.8 \\
        \hline
        Ours & 56.9 $\pm$ 0.3 & 50.6 $\pm$ 0.1 & 72.0 $\pm$ 0.3 & 73.6 $\pm$ 0.3 & 63.3\\     
         \hline
      \end{tabular}
  }
  \label{tab:oh}
  \end{table*}

\begin{table*}[htbp]
  \renewcommand\arraystretch{1.1}
    \centering
    \caption{
      Average accuracy results on the TerraIncognita~\cite{beery2018recognition} dataset according to the DomainBed benchmark~\cite{gulrajani2020search}.
      }
    \adjustbox{max width=\textwidth}{%
    \begin{tabular}{lccccc}
        \hline 
        \textbf{Method} & \textbf{L100} & \textbf{L38} & \textbf{L43} & \textbf{L46} & \textbf{Avg.} \\
        \hline 
        HYPO & 43.2  & 32.5 & 32.4 &  37.2 & 36.3 \\
        DecAug & 39.7 & 39.7 &  56.8 & 36.2 & 43.1 \\
        VNE & 49.8$\pm$ 1.3 & 33.4 $\pm$  1.6 & 52.3 $\pm$  0.5 & 34.6 $\pm$  0.7 & 42.5\\
        ERM &42.1 $\pm$ 2.5 &30.1 $\pm$ 1.2 &48.9 $\pm$ 0.6 &34.0 $\pm$ 1.1 &38.8 \\
        IRM &41.8 $\pm$ 1.8 &29.0 $\pm$ 3.6 &49.6 $\pm$ 2.1 &33.1 $\pm$ 1.5 &38.4 \\
        Mixup &49.4 $\pm$ 2.0 &35.9 $\pm$ 1.8 &53.0 $\pm$ 0.7 &30.0 $\pm$ 0.9 &42.1 \\
        MLDG &39.6 $\pm$ 2.3 &33.2 $\pm$ 2.7 &52.4 $\pm$ 0.5 &35.1 $\pm$ 1.5 &40.1 \\
        CORAL &46.7 $\pm$ 3.2 &36.9 $\pm$ 4.3 &49.5 $\pm$ 1.9 &32.5 $\pm$ 0.7 &41.4 \\
        MMD &49.1 $\pm$ 1.2 &36.4 $\pm$ 4.8 &50.4 $\pm$ 2.1 &32.3 $\pm$ 1.5 &42.0 \\
        DANN &44.3 $\pm$ 3.6 &28.0 $\pm$ 1.5 &47.9 $\pm$ 1.0 &31.3 $\pm$ 0.6 &37.9 \\
        CDANN &36.9 $\pm$ 6.4 &32.7 $\pm$ 6.2 &51.1 $\pm$ 1.3 &33.5 $\pm$ 0.5 &38.6 \\
        MTL &45.2 $\pm$ 2.6 &31.0 $\pm$ 1.6 &50.6 $\pm$ 1.1 &34.9 $\pm$ 0.4 &40.4 \\
        SagNet &36.3 $\pm$ 4.7 & 40.3 $\pm$ 2.0 &52.5 $\pm$ 0.6 &33.3 $\pm$ 1.3 &40.6 \\
        ARM &41.5 $\pm$ 4.5 &27.7 $\pm$ 2.4 &50.9 $\pm$ 1.0 &29.6 $\pm$ 1.5 &37.4 \\
        VREx &48.0 $\pm$ 1.7 &41.1 $\pm$ 1.5 &51.8 $\pm$ 1.5 &32.0 $\pm$ 1.2 &43.2 \\
        RSC &42.8 $\pm$ 2.4 &32.2 $\pm$ 3.8 &49.6 $\pm$ 0.9 &32.9 $\pm$ 1.2 &39.4 \\
        SelfReg &46.1 $\pm$ 1.5 &34.5 $\pm$ 1.6 &49.8 $\pm$ 0.3 &34.7 $\pm$ 1.5 &41.3 \\
        GroupGRO &45.3 $\pm$ 4.6 &36.1 $\pm$ 4.4 &51.0 $\pm$ 0.8 &33.7 $\pm$ 0.9 &41.5 \\
        MixStyle &50.6 $\pm$ 1.9 &28.0 $\pm$ 4.5& 52.1$\pm$0.7 &33.0 $\pm$ 0.2 &40.9 \\
        Fish &46.3 $\pm$ 3.0 &29.0 $\pm$ 1.1 &52.7 $\pm$ 1.2 &32.8 $\pm$ 1.0 &40.2 \\
        SD &45.5 $\pm$ 1.9 &33.2 $\pm$ 3.1 &52.9 $\pm$ 0.7 &36.4 $\pm$ 0.8 &42.0 \\
        CondCAD &44.4 $\pm$ 2.9 &32.9 $\pm$ 2.5 &50.5 $\pm$ 1.3 &30.8 $\pm$ 0.5 &39.7 \\
        Fishr   &49.9 $\pm$ 3.3 &36.6 $\pm$ 0.9 &49.8 $\pm$ 0.2 &34.2 $\pm$ 1.3 &42.6 \\
        MIRO &46.0 $\pm$ 0.7 &34.4 $\pm$ 0.4 &51.2 $\pm$ 1.0 &33.6 $\pm$ 0.9 &41.3 \\
        DGRI  &46.2 $\pm$ 4.0 &39.7 $\pm$ 2.4 &53.0 $\pm$ 0.6 &36.0 $\pm$ 0.3 &43.7 \\
        W2D & 34.3  $\pm$ 4.3 & 27.3 $\pm$ 2.1 & 47.4 $\pm$ 0.4 & 34.4 $\pm$ 1.3 & 35.9 \\
        \hline
        Ours & 46.9 $\pm$ 1.9 & 38.3 $\pm$ 1.5 & 54.5 $\pm$ 1.1 & 37.5 $\pm$ 0.9 & 44.3 \\
        \hline
        \end{tabular}
  }
  \label{tab:tr}
  \end{table*}

\end{document}